\documentclass{article} 
\usepackage{iclr2027_conference,times}

\usepackage{amsmath,amsfonts,bm}

\def\eqref#1{equation~\ref{#1}}

\def\1{\bm{1}}

\DeclareMathAlphabet{\mathsfit}{\encodingdefault}{\sfdefault}{m}{sl}
\SetMathAlphabet{\mathsfit}{bold}{\encodingdefault}{\sfdefault}{bx}{n}

\usepackage{booktabs}
\usepackage{amsmath}
\usepackage{amssymb}
\usepackage{amsthm}         
\usepackage{microtype}
\usepackage{natbib}
\usepackage{enumitem}
\usepackage{graphicx}
\usepackage{multirow}
\usepackage{wrapfig}

\theoremstyle{plain}

\usepackage{hyperref}
\usepackage{url}

\title{When Does Correction Become Repair? Mechanistic Auditing of Internal Interventions in
Tool-Using LLMs}

\author{%
  Jiayi Li $^{1}$
  \quad
  Ruizhe Li$^{2\thanks{Corresponding Author: \texttt{r.li.7@bham.ac.uk}}}$
\\
  $^1$University of the Chinese Academy of Sciences, China \\
  $^2$School of Computer Science, University of Birmingham, UK
}

\iclrfinalcopy 
\begin{document}

\maketitle
\begin{abstract}
Before invoking external tools, an agentic LLM must select among a $K$-way action space: executing a call, seeking clarification, answering directly, or declining. While internal activation steering can alter these pre-execution decisions, conventional aggregate metrics obscure where altered states land and what collateral damage they inflict. We present \textbf{SAKIKO}, an auditing framework that formalizes representation repair via directional error discovery, router-conditioned intervention, destination-resolved verification, and prospectively frozen statistical licensing. Across seven LLMs on When2Call and MetaTool, channel-keyed interventions induce direction-specific net gains in five models; across three sealed evaluations, none of 59 budget-matched random directions matches calibrated target gain. Crucially, destination auditing shows that behavioral movement does not equal repair: an intervention achieving $+55$ net gain corrupts over half of the baseline-correct decisions it touches, and promising point estimates on Qwen3-4B and Gemma-2-9B are formally declined due to finite-sample uncertainty. \textbf{SAKIKO} establishes the necessity of outcome-resolved adjudication before claiming internal repair.
\end{abstract}

\section{Introduction}
\label{sec:intro}

The evolution of LLMs into autonomous agents fundamentally alters how model failures arise. Beyond factual hallucination in generated text~\citep{ji2023survey}, tool-augmented agents face a structurally distinct failure mode: \textit{pre-execution tool-decision errors}. Before an external API is invoked, evidence retrieved, or execution feedback received, an agent must decide upon an appropriate course of action. It may invoke a tool when prerequisites are absent, answer prematurely when clarification is required, decline an answerable query, or omit an indispensable external tool call. In these scenarios, the failure is situated not in the generated content, but in the upstream selection of action type.

While extensive research has advanced downstream execution, such as API parameter synthesis, multi-tool composition, and retrieval~\citep{schick2023toolformer, tang2023toolalpaca, patil2024gorilla, qin2024toolllm}, the upstream decision of \textit{whether} and \textit{when} to act~\citep{wei2022emergent, qin2024tool} remains underexplored as a target for representation-level intervention. Recent mechanistic studies, such as the Activation Steering Adapter (ASA)~\citep{wang2026asa} and CAST~\citep{lee2025programming}, demonstrate that pre-execution states can be conditionally decoded and guided via internal activations. However, existing interventions typically assume binary action spaces (e.g., tool vs.\ no-tool), where departing an incorrect state deterministically lands the model in the correct one. In realistic agentic environments with $K$-way action topologies, leaving an erroneous state provides no guarantee of reaching the target, and an intervention may simply displace probability mass to a third, equally invalid action. Consequently, \textbf{behavioral movement does not constitute repair}.

This distinction motivates our central question: \textit{When an internal intervention alters a pre-execution tool decision, when does that change constitute genuine repair rather than mere behavioral movement?} We approach this problem through a scientific progression across three core dimensions: i) \textbf{Correction}: Can directional error channels be discovered from a model's internal representation, detected at inference, and selectively steered? ii) \textbf{Verification}: Does the intervention transition the model to the valid target action, and does it preserve decisions that were already correct? iii) \textbf{Licensing}: Do the observed destination outcomes and their quantified uncertainty provide sufficient empirical evidence to license a formal claim of repair?

\begin{figure}[tb]
    \centering
    \vspace{-1.5em}
    \includegraphics[width=\linewidth]{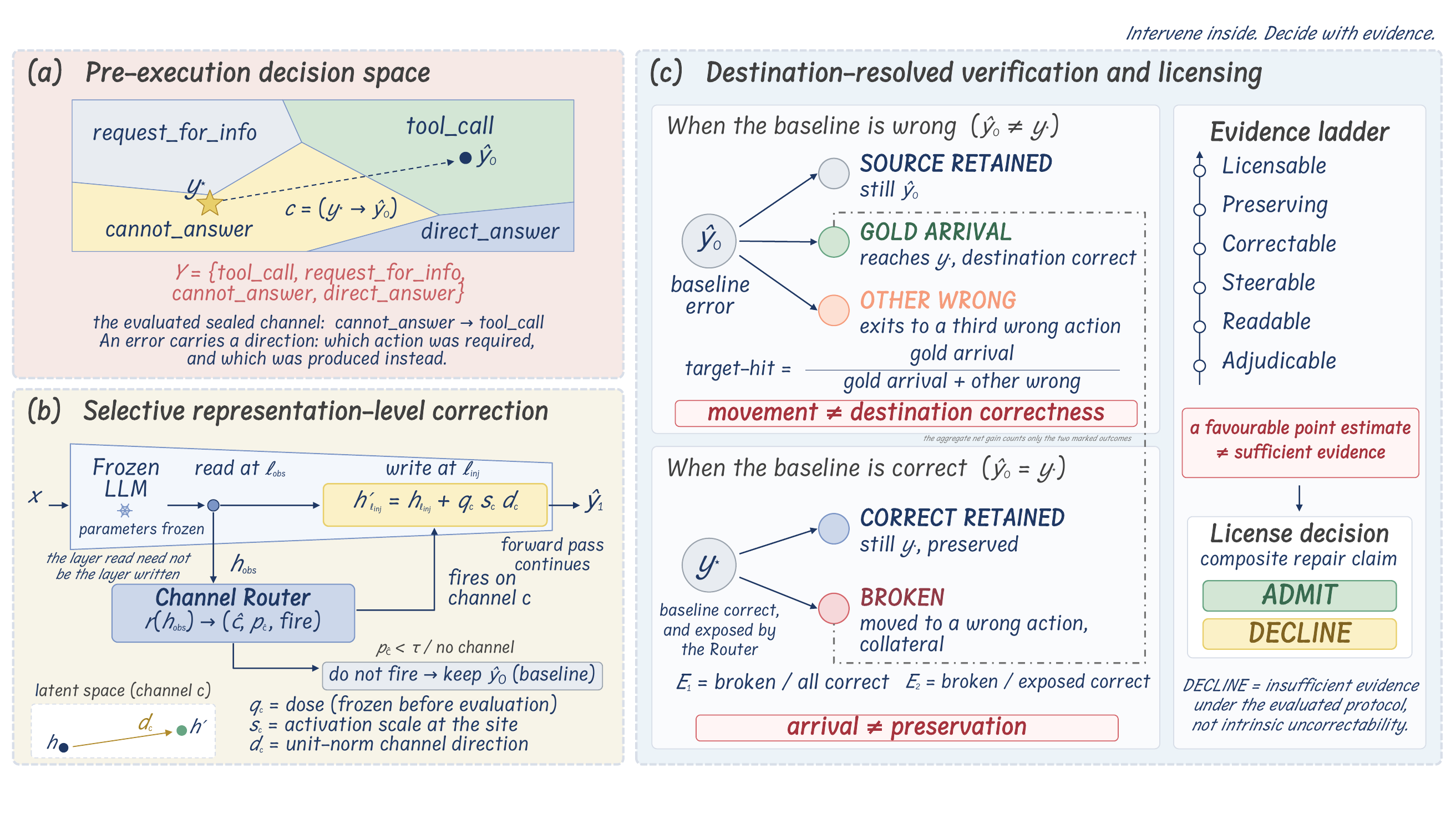}
    \vspace{-.5in}
    \caption{
        Overview of SAKIKO. The framework discovers directional pre-execution tool-decision error channels, detects channel-associated states per input, applies the corresponding channel-keyed intervention, verifies the post-intervention destination and collateral on initially correct decisions, and licenses only the strongest repair claim supported by the evidence.
    }
    \label{fig:introcore}
    \vspace{-1.5em}
\end{figure}

To operationalize this progression, we introduce \textbf{SAKIKO} (\textbf{S}tructured \textbf{A}djudication of \textbf{K}eyed \textbf{I}nterventions over \textbf{K}-way \textbf{O}utcomes), a staged adjudication framework that treats representation-level repair as an evidence process rather than an isolated steering operation (Fig.~\ref{fig:introcore}). SAKIKO structures intervention and evaluation across five functional stages, i.e., \textbf{Discovery} of a model's directional error channels, \textbf{Detection} of channel-associated states per input, \textbf{Keyed Intervention} applied only where a channel fires, \textbf{Destination-Resolved Verification} over the complete $K$-way action space, and \textbf{Statistical Licensing} against a prospectively frozen evidence hierarchy.

Our empirical evaluations demonstrate why each stage of this pipeline does distinct work. At the \textit{Correction} stage, channel-keyed interventions succeed in altering behavior, producing marked net gains (e.g., $+79$ on the locked test set for Qwen2.5-7B and consistent positive gains across all seeds for Phi-3.5). 
Subjecting those corrections to \textit{Verification} and \textit{Licensing} uncovers three dissociations that aggregate metrics conceal. A Phi-3.5 intervention worth a net $+55$ corrupts 52 of the 93 baseline-correct decisions its Router fires on. On Qwen3-8B, activation steering and an output-score baseline reach near-identical net gains at target-hit rates of $0.7308$ against $0.6379$, so one lands on the intended target while the other redistributes mass across alternate errors. And Qwen3-4B and Gemma-2-9B carry promising point estimates ($0.5781$ and $0.6296$) that \textsc{SAKIKO} declines to license, because their confidence bounds at the available sample sizes remain inconclusive.
These empirical observations confirm that \textit{behavioral improvement, destination-correct repair, preservation of correct behavior, and evidential sufficiency do not inherently coincide}.

Our main contributions are as follows:
\begin{enumerate}[leftmargin=*,nosep]
    \item \textbf{Conceptual Framing}: We formalize pre-execution tool-decision errors within $K$-way action spaces, demonstrating that leaving an error state cannot be equated with target repair.
    \item \textbf{SAKIKO Framework}: We formulate an end-to-end pipeline that integrates directional error discovery, conditioned steering, destination-resolved verification, and statistical licensing.
    \item \textbf{Empirical Adjudication \& Controls}: Across seven LLMs spanning five model families and three benchmarks in three roles, with an eighth model stopped at pre-intervention screening, we test channel-keyed steering against structural controls
    to show that headline accuracy metrics can conceal collateral damage and destination redistribution.
    \item \textbf{Mechanistic Decoupling}: We show that linear decodability does not imply steerability, and optimal injection site, dosage, and intervention success diverge from probe accuracy, showing that verification cannot be reduced to internal detection alone.
\end{enumerate}

\section{Related Work}
\label{sec:related_work}

\textbf{Tool Learning and Epistemic Selection.}
Prior tool learning focuses on downstream execution: API adherence, parameter synthesis and environment generalization via fine-tuning or retrieval~\citep{schick2023toolformer,qin2024toolllm,tang2023toolalpaca,patil2024gorilla,qin2024tool}. Before parameterization, however, an agent must resolve the epistemic decision of \textit{whether} external action is needed. \citet{wang2026positionagentinvokeexternal} taxonomize failure modes such as tool bypass and over-delegation, but treat the agent as a black box, omitting latent geometry and representation-level intervention. Closest to our setting, \citet{wu2026call} read tool need from hidden states to drive controllers, but score the outcome by task performance rather than by where redirected decisions land.

\textbf{Agent Observability and Activation Steering.}
Pre-execution states are linearly decodable, enabling real-time detection of tool hallucinations by residual-stream probing~\citep{healy2026internalrepresentationsindicatorshallucinations} and localized risk monitoring by sparse autoencoders~\citep{tatsat2026blackboxinterpretabilityagentic}. Beyond passive detection, representation engineering~\citep{zou2025representationengineeringtopdownapproach,lee2025programming} enables direct control: linear steering vectors invert tool choices in constrained menus~\citep{wu2026tool}, and router-conditioned vectors overcome inertia in multi-turn settings~\citep{wang2026asa}. Closest in apparatus, \citet{shi2026overcalling} recover a sparse-autoencoder basis for the call/no-call decision on the benchmark we also use, estimate an activation-independent CALL offset, and cancel it with a closed-form counter-bias shift.

\textbf{SAKIKO Distinction: Audited Repair in $K$-Way Action Spaces.}
Prior tool steering evaluates narrow binary exits or pairwise tool swaps~\citep{wu2026tool,shi2026overcalling}, while diagnostic multiclass work remains observational~\citep{zhao2026calibration}. Crucially, behavioral readouts often diverge from internal representation effects~\citep{jiang2026behavioural}. In $K$-way action topologies, simply measuring state exits obscures destination divergence, collateral degradation, and lack of statistical support. Moving beyond passive probes~\citep{healy2026internalrepresentationsindicatorshallucinations,zhao2026calibration} and uncalibrated steering~\citep{wu2026tool,wang2026asa}, \textsc{SAKIKO} establishes an end-to-end adjudication pipeline: discovering channel vectors, verifying multiclass destinations, auditing baseline preservation, and enforcing preregistered uncertainty bounds for certified repair (App.~\ref{app:related_work} for more discussion).


\section{The SAKIKO Framework}
\label{sec:framework}
\textsc{SAKIKO} formalizes the criteria under which an internal intervention on a pre-execution tool decision constitutes a valid claim of repair. Rather than proposing another ad-hoc steering vector, \textsc{SAKIKO} frames evaluation as a non-collapsing evidentiary pipeline: \textit{Correction} $\rightarrow$ \textit{Destination Correctness} $\rightarrow$ \textit{Preservation} $\rightarrow$ \textit{Evidential Sufficiency} $\rightarrow$ \textit{Licensed Repair}. In multiclass action spaces, no property guarantees the next (Figs.~\ref{fig:thesis} and~\ref{fig:destination}): vacating an error can divert mass to another invalid action, the same intervention can corrupt decisions that were already correct, and positive point estimates may lack statistical significance. Consequently, \textsc{SAKIKO} decouples representation steering~\citep{lee2025programming, wang2026asa} from destination verification and formal statistical licensing.

\subsection{Problem Formulation and Outcome Topology}
\label{subsec:problem}

Let $\mathcal{Y}$ be the pre-execution action space (direct response, API call, clarification request, task declination). For an input $x$ a frozen agent produces a baseline action $\hat{y}_0 \in \mathcal{Y}$, evaluated against a reference $y^\star \in \mathcal{Y}$, which partitions inputs into baseline-correct and baseline-error populations. Rather than treating errors monolithically we model failure as an ordered \textbf{directional error channel} $c \equiv (y^\star \rightarrow \hat{y}_0)$, with gold target $g \equiv y^\star$ and erroneous source $s \equiv \hat{y}_0$ ($g \neq s$). The intervention emits an updated decision $\hat{y}_1 \in \mathcal{Y}$. This induces an exhaustive, mutually exclusive five-way partition (Table~\ref{tab:outcomes}, App.~\ref{app:notation}): the $N_c$ channel errors split into \textsc{source retained} ($S_c$), \textsc{gold arrival} ($A_c$) and \textsc{other wrong} ($O_c$), with source exits $X_c = A_c + O_c$, and the baseline-correct decisions the Router touches ($C_{\mathrm{exp}}$) split into \textsc{correct retained} and \textsc{broken} ($B$). Conventional benchmarks assess steering via an aggregate net gain, and two conventions are in use. We keep them apart throughout: $G_{\mathrm{whole}} = \mathrm{Fixed} - B$ counts every baseline error in the evaluation population that becomes correct, while $G_{\mathrm{channel}} = \sum_c A_c - B$ counts only arrivals inside the adjudicated channel, so the first exceeds the second whenever correction reaches an error the channel does not key ($+40$ against $+38$ on Qwen3-8B). Every net quantity we report carries one of the two labels. However, the map from the five outcome classes to either aggregate is
many-to-one: a given net gain does not determine $\mathrm{TH}_c$ or the
collateral, as App.~\ref{app:notation} shows.

\subsection{Overview of the SAKIKO Pipeline}
\label{subsec:overview}

\begin{figure}[t]
    \centering
    \includegraphics[width=\linewidth]{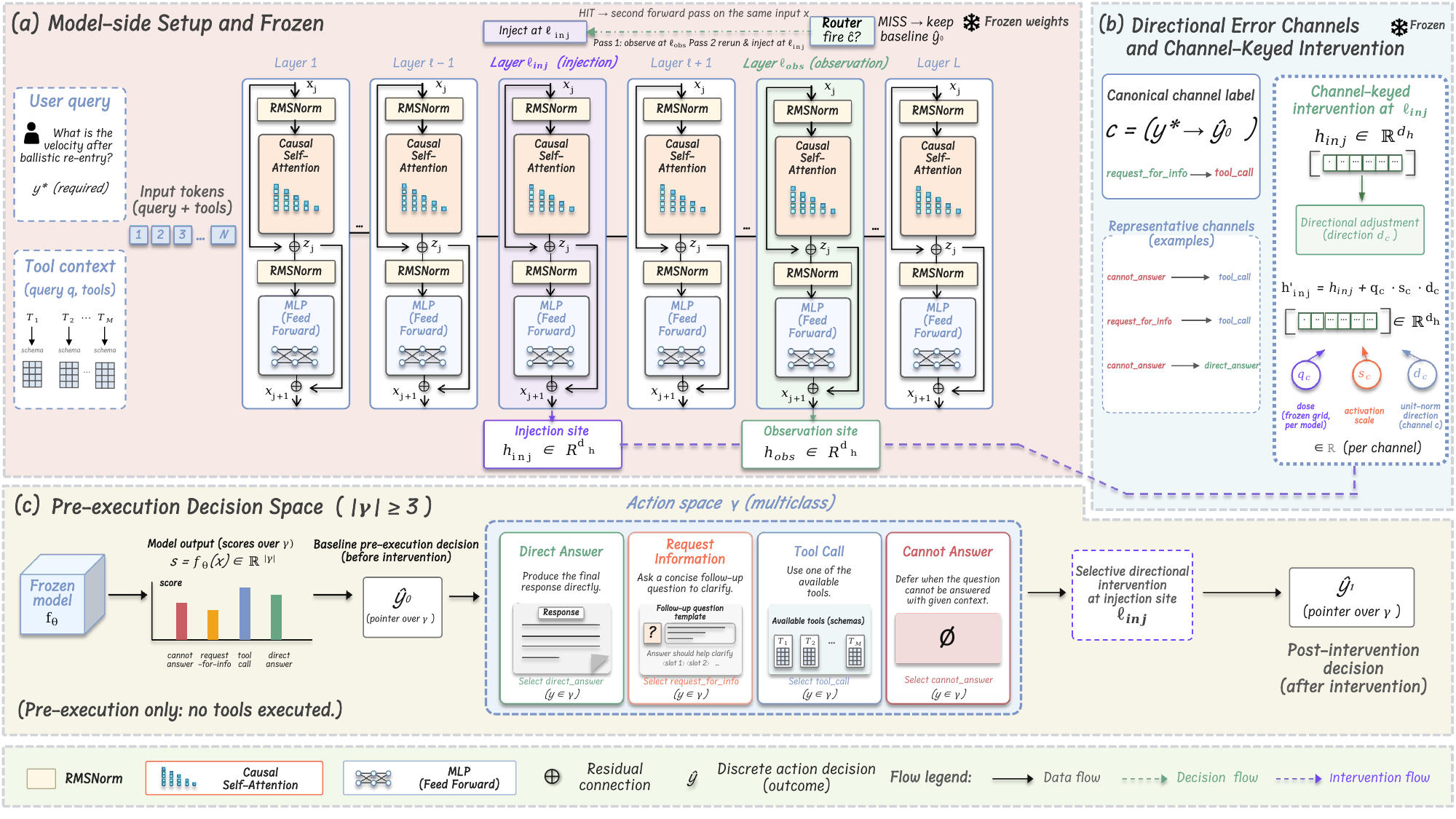}
    \vspace{-.3in}
    \caption{\textbf{Model-side view of SAKIKO.} \textbf{(a)} Over a frozen model, a
    first pass reads hidden state at observation layer $\ell_{\mathrm{obs}}$
    and a channel Router decides whether to fire; on a miss the baseline decision
    $\hat{y}_0$ stands, and on a hit the input is rerun so the correction can be
    applied at the injection layer $\ell_{\mathrm{inj}}$, which precedes
    $\ell_{\mathrm{obs}}$ in the stack. \textbf{(b)} Each baseline error is keyed by
    its direction $c = (y^\star \rightarrow \hat{y}_0)$ and corrected by adding
    $q_c s_c d_c$ at that site, with $d_c$ unit-norm, $s_c$ a local activation scale
    and $q_c$ a dose frozen before evaluation. \textbf{(c)} The decision precedes any
    tool call: a frozen readout scores the action space $\mathcal{Y}$ and its argmax
    gives $\hat{y}_0$, and $\hat{y}_1$ after intervention, with destination,
    collateral and licensing.}
    \vspace{-1em}
    \label{fig:framework}
\end{figure}

\textsc{SAKIKO} introduces a five-stage pipeline integrating representation-level steering with statistical auditing (Fig.~\ref{fig:framework}):
(1)~\textit{Discovery} isolates directional error channels $(g \rightarrow s)$ from baseline confusion matrices;
(2)~\textit{Detection} fits a linear Router over frozen mid-layer representations, with a firing threshold selected on validation data, to gate interventions;
(3)~\textit{Correction} steers routed inputs along channel-specific latent vectors;
(4)~\textit{Destination Verification} tracks post-intervention trajectories across the full $K$-way action topology for both error and clean cohorts; and
(5)~\textit{Statistical Licensing} bounds empirical gains via a pre-registered evidence hierarchy, certifying repair only when confidence bounds rule out stochastic drift.
Therefore, while latent manipulation confirms steerability, only structured adjudication certifies genuine repair.

\label{subsec:detect}
\textbf{Channel Discovery and Latent Detection.} Rather than assuming error topologies \textit{a priori}, \textsc{SAKIKO} enumerates them from the training-split confusion matrix and retains directional channels $c = (y^\star \rightarrow \hat{y}_0)$ that clear a support minimum on both the training and validation splits (App.~\ref{app:discover} and~\ref{app:support}). Because dominant error modes vary across architectures and share non-trivial cosine similarities, e.g., on Qwen2.5-7B's three When2Call channels, pairwise cosines of $0.488$, $0.569$ and $0.706$ measured at a common layer, a single monolithic tool-error vector conflates functionally disparate failures. For each retained channel $c$, we train a linear Router $r(h_{\mathrm{obs}}) \rightarrow (\hat{c}, p_{\hat{c}})$ on final-token representations $h_{\mathrm{obs}}$ at layer $\ell_{\mathrm{obs}}$, triggering steering iff $p_{\hat{c}} \ge \tau_{\hat{c}}$, and unrouted inputs remain unperturbed ($\hat{y}_1 = \hat{y}_0$). Crucially, probe decodability does not guarantee steerability: while Router accuracy plateaus across intermediate layers, downstream steering efficacy drops sharply (\S~\ref{sec:mechanism}). The Router thus serves to bound exposure on baseline-correct samples ($C_{\mathrm{exp}}$), not to guarantee repair.

\textbf{Channel-Keyed Intervention Protocol.} Upon Router activation for channel $c$, we intervene at layer $\ell_{\mathrm{inj}}$ on the MLP output, before it rejoins the decoder residual stream, at every sequence position, parameterized by a fixed tuple $\theta_c = \{d_c, \ell_{\mathrm{obs}}, \ell_{\mathrm{inj}}, q_c, \tau_c\}$:
$h'_{\ell_{\mathrm{inj}}} = h_{\ell_{\mathrm{inj}}} + q_c \, s_c \, d_c$,
where $d_c$ is a unit-norm steering direction, $s_c$ is a layer scale factor, and $q_c$ denotes normalized dosage. Subsequent decoding remains unmodified. If $\ell_{\mathrm{inj}} \le \ell_{\mathrm{obs}}$, two-pass inference avoids lookahead leakage. Structural controls run as each protocol permits: budget-matched random vectors everywhere, sign reversals and layer misallocations where the records permit, and cross-channel substitution only in the historical battery, which the frozen sealed inventory excludes by construction (App.~\ref{app:controls}).

\textbf{Destination-Resolved Verification and Collateral Auditing.} Verification resolves exact transitions across $\mathcal{Y}$. On baseline errors, destination correctness is read over source exits as $\mathrm{TH}_c = A_c / X_c$, separating gold arrivals from lateral redistribution into $O_c$, and as $\mathrm{TG}_c = (A_c - O_c) / N_c$, which normalises over the whole channel and so carries coverage as well as composition. Collateral is reported on two denominators, $\mathrm{E1} = B / |C|$ and $\mathrm{E2} = B / |C_{\mathrm{exp}}|$ with $\mathrm{E2} \ge \mathrm{E1}$: because a selective Router minimises $|C_{\mathrm{exp}}|$, $\mathrm{E1}$ alone understates the risk to a touched decision by $|C| / |C_{\mathrm{exp}}|$. Appendix~\ref{app:notation} defines every quantity against its named population.

\label{subsec:license}
\textbf{Statistical Licensing and Repair Adjudication.} Licensing evaluates empirical evidence against a pre-registered evidence hierarchy (Table~\ref{tab:ladder}, App.~\ref{app:licensing}), assessing evidentiary rigor rather than raw model capability. A configuration $\pi$ earns a formal repair claim iff it satisfies the conjunction, in which $\textsc{PRESERVING}_{\text{pop}}$ is the population-level form the frozen gate tests and not the stronger exposure-conditional one (App.~\ref{app:licensing}):
$ \textsc{REPAIR}(\pi) \;\equiv\; 
    \textsc{CORRECTABLE}(\pi) \;\wedge\; 
    \textsc{PRESERVING}_{\text{pop}}(\pi) \;\wedge\; 
    \textsc{LICENSABLE}(\pi)$.
Under our protocol, adjudication is governed by a ten-condition gate evaluating sample support, specificity over matched-random baselines ($p \leq 0.05$), threshold criteria for $\mathrm{TH}_c$ and $\mathrm{E1}$, bootstrap confidence bounds, and exact baseline replication (\S~\ref{sec:licensing}). The gate outputs an explicit verdict: \textsc{ADMIT} licenses the repair claim for that setting under the frozen conditions, while \textsc{DECLINE} rejects it and isolates the failed condition. The licence is narrower than the property it is named for. Preservation is adjudicated on $\mathrm{E1}$, the denominator the preregistration fixed, so an \textsc{ADMIT} certifies that collateral damage is bounded across the correct population and \emph{not} that it is bounded on the decisions the Router actually touches; no setting we evaluate attains the latter (\S~\ref{subsec:preservation}). By separating unverified steering from a licensed claim, \textsc{SAKIKO} prevents aggregate net gains or optimistic point estimates from obscuring localized intervention failures.

\section{Empirical Evaluation: Channel-Keyed Correction}
\label{sec:results}

We evaluate the first stage of the \textsc{SAKIKO} progression: whether channel-keyed internal interventions produce genuine, direction-specific corrections of pre-execution tool decisions. We address three sequential questions: (1) Does intervening along an extracted channel direction shift decisions toward the target action? (2) Is this displacement driven by the geometry of the identified direction rather than arbitrary representation noise? (3) How reliably does this steerability hold across diverse model architectures? We confine this analysis to the \textit{Correction} stage, and destination-resolved outcomes, collateral costs on baseline-correct samples, and statistical licensing are evaluated in \S~\ref{sec:correction_to_repair}.

\subsection{Experimental Setup}
\label{sec:setup}

\textbf{Tasks and Readout Protocol.}
We treat three benchmarks independently rather than pooling results (App.~\ref{app:data}). We primarily evaluate on When2Call~\citep{ross2025when2call}, scoring its four pre-execution actions via teacher forcing to cleanly isolate steering effects. We test binary bidirectional steering using MetaTool~\citep{huang2024metatool} on Qwen2.5-7B. ACEBench provides an out-of-domain four-action ontology evaluated via free generation and rule parsing. Although the readout protocol certified successfully ($0.756$ accuracy; $0.687$ macro-F1), ACEBench failed downstream channel discovery criteria, i.e., specifically lacking minimum transition support and yielding sub-threshold paraphrase agreement ($0.56$ vs.\ $0.80$). Per our preregistered protocol, ACEBench was retired without intervention experiments. This negative case demonstrates that our framework can rigorously validate readout stability and screen out unviable task topologies prior to intervention.

\textbf{Models, Protocols, and Metrics.} We evaluate seven LLMs (3.8B--9B) across two protocol cohorts using a 2,556/548/548 train/val/test split. The \textit{historical} cohort (Phi-3.5-mini~\citep{abdin2024phi3}, Qwen2.5-7B~\citep{yang2024qwen2}, Llama-3.1-8B~\citep{grattafiori2024llama3}, and Mistral-7B~\citep{jiang2023mistral}) targets three channels, which on Qwen2.5-7B cover about $78\%$ of errors, evaluated via whole-population net gain $G_{\mathrm{whole}}$. The prospective \textit{sealed} cohort (Qwen3-4B, Qwen3-8B~\citep{yang2025qwen3}, and Gemma-2-9B~\citep{gemmateam2024gemma2}) pre-registers all vectors, thresholds, and doses $\in \{0.125, 0.25, 1.0\}$ on the \texttt{cannot\_answer}~$\rightarrow$~\texttt{tool\_call} channel prior to test unblinding, evaluated via target gain $\mathrm{TG}_c = (A_c - O_c) / N_c$. Each run is benchmarked against the structural controls its protocol permits: zero, budget-matched random, reversed and wrong-layer arms in the sealed inventory, with cross-channel substitution excluded from that inventory by construction and run only on Phi-3.5-mini (App.~\ref{app:controls}); each sealed setting executes $65$ arms in total, six named and $59$ random. Sealed significance is adjudicated against those 59 directions by an add-one Monte Carlo test over the empirical null, $p = (1 + \sum \mathbb{I}[\mathrm{TG}_{\mathrm{rand}} \ge \mathrm{TG}_{\mathrm{real}}]) / 60$; they are isotropic unit vectors rescaled to the calibrated budget $q_c s_c$, drawn under seeds disjoint from development and hash-pinned before the split was opened (App.~\ref{app:controls}).

\begin{figure}[tb]
    \centering
    \vspace{-1em}
    \includegraphics[width=0.88\linewidth]{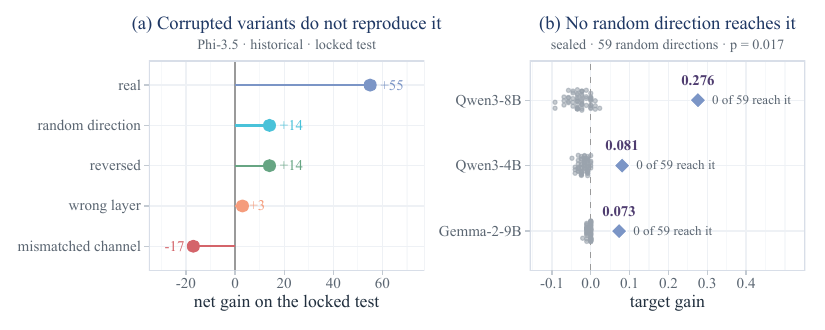}
    \vspace{-.25in}
    \caption{\textbf{Correction is direction-specific.} \textbf{(a)} On Phi-3.5 locked test, calibrated direction yields a net gain of $+55$, whereas random and sign-reversed vectors yield $+14$, layer shifts yield $+3$, and mismatched channel vectors incur $-17$ (arms colored by corrupted component; four arms show aggregate counts). \textbf{(b)} Under sealed protocol, all 59 budget-matched random vectors passing the Router gate trail calibrated target gain (add-one $p = 0.017$ across all settings).}
    \vspace{-1.5em}
    \label{fig:specificity}
\end{figure}

\subsection{Internal Interventions Induce Targeted Action Transitions}
\label{subsec:correction_occurs}

\textbf{Quantitative Gains Across Architectures and Task Formulations.}
Channel-keyed interventions consistently drive intended behavioral shifts across model families and tasks. In the historical cohort on When2Call, intervention yields net gains of $+79$ decisions on Qwen2.5-7B (92 corrections vs.\ 13 regressions) and $+55$ on Phi-3.5-mini. In the sealed cohort, all models achieve positive target gains ($\mathrm{TG}_c$) on the frozen \texttt{cannot\_answer}~$\rightarrow$~\texttt{tool\_call} channel: Qwen3-8B reaches $\mathrm{TG}_c = 0.276$ ($+24$ net arrivals / $87$ errors), Qwen3-4B achieves $\mathrm{TG}_c = 0.081$ ($+10 / 124$), and Gemma-2-9B achieves $\mathrm{TG}_c = 0.073$ ($+7 / 96$). On Qwen3-8B, the Router fires on $82.8\%$ of errors, triggering $59.8\%$ source exits and $43.7\%$ target arrivals. Finally, bidirectional steering on MetaTool with Qwen2.5-7B corrects both premature and omitted tool calls, yielding mean net gains of $+8.6$ and $+13.0$ across five independent runs.

\textbf{Localized Latent Steerability vs. Global Propensity Drift.}
These results confirm that pre-execution tool decisions can be selectively steered in latent space at inference time. Crucially, bidirectional corrections on opposing MetaTool channels demonstrate that interventions act via channel-keyed rectification rather than indiscriminate shifts in global tool-use propensity. However, while target gains confirm departure from source errors across binary and 4-way settings, aggregate displacement does not ensure ground-truth arrival, which motivates \textsc{SAKIKO}'s destination verification stage.

\subsection{Empirical Confirmation of Directional Specificity}
\label{subsec:specificity}

\textbf{Adjudication Against Structural Counterfactuals and Permutations.}
Rigorous structural controls confirm that behavioral shifts stem from specific channel geometry rather than arbitrary perturbation. On Phi-3.5-mini (Fig.~\ref{fig:specificity}), calibrated vector ($+55$ net gain) dramatically outperforms norm-matched random ($+14$), sign-reversed ($+14$), layer-misallocated ($+3$), and performance-degrading cross-channel baselines. Likewise, on Qwen2.5-7B, calibrated direction ($+79$) substantially exceeds sign-reversed controls ($+16$) and ten budget-matched random vectors (mean $+25.2$, max $+50$). In sealed evaluations, calibrated intervention decisively beats the empirical null of 59 pre-registered random vectors, achieving the minimum add-one Monte Carlo significance floor ($p = 1/60 = 0.017$) across all architectures. Calibrated vectors yield significantly more gold arrivals than random baselines on Qwen3-8B ($38$ vs.\ max $8$), Qwen3-4B ($37$ vs.\ $2$), and Gemma-2-9B ($17$ vs.\ $0$), while zero-dose, reversed, and wrong-layer perturbations cause negligible displacement ($\mathrm{TG}_c \le 0.058$).

\textbf{Geometric Alignment and Layer-Specific Causal Control.}
These counterfactual adjudications confirm that behavioral repair is causally governed by internal directional alignment. The failure of layer-mismatched and cross-channel controls demonstrates that intervention requires dual specificity: the vector must encode the exact channel geometry and target the precise layer arbitrating tool selection. Because uncalibrated noise and arbitrary perturbations fail to produce targeted gold arrivals, the intervention acts via site- and direction-specific control rather than a non-specific activation shock, though it identifies locus of intervention rather than fine-grained circuit components.

\subsection{Cross-Model Analysis and Pre-Intervention Attrition}
\label{subsec:cross_model}

\textbf{Model Screening, Structural Failures, and Attrition Statistics.}
Directional specificity is confirmed in five of the seven evaluated architectures, alongside two control failures and one pre-intervention screening exit (Table~\ref{tab:correction_summary}); together with the retired third benchmark, this demonstrates that the protocol halts unviable settings as rigorously as it licenses valid ones. In the historical cohort, Llama-3.1-8B and Mistral-7B fail structural specificity: for Llama-3.1-8B, 5 of 20 random vectors match or exceed intervention gains, while on Mistral-7B, the calibrated direction ($+12$) underperforms both sign-reversed ($+19$) and mean random baselines ($+29$). In the prospective cohort, despite the highest baseline accuracy ($0.4533$), Qwen3.5-9B was disqualified before intervention due to severe class skew, i.e., favoring information requests ($0.326$) over refusals ($0.087$). This produced ample training/validation errors (300/170) but only 29 baseline-correct reference instances, violating the preregistered minimum support threshold of 30 and triggering an automatic protocol halt.

\textbf{Non-Universality of Linear Steering and Role of Protocol Gates.}
Linear steerability is neither universally shared across architectures nor guaranteed by high baseline accuracy. In models like Mistral-7B and Llama-3.1-8B, latent steering behaves as non-specific noise, inducing stochastic drift rather than targeted repair. Furthermore, the pre-intervention disqualification of Qwen3.5-9B demonstrates the critical utility of \textsc{SAKIKO}'s preregistered gates: by filtering out under-supported reference populations prior to unblinding, the protocol prevents underpowered point estimates from producing false claims of repair before verification begins.

\section{Analysis: From Directional Correction to Licensed Repair}
\label{sec:correction_to_repair}

\begin{figure}[t]
    \centering
    \vspace{-1em}
    \includegraphics[width=0.88\linewidth]{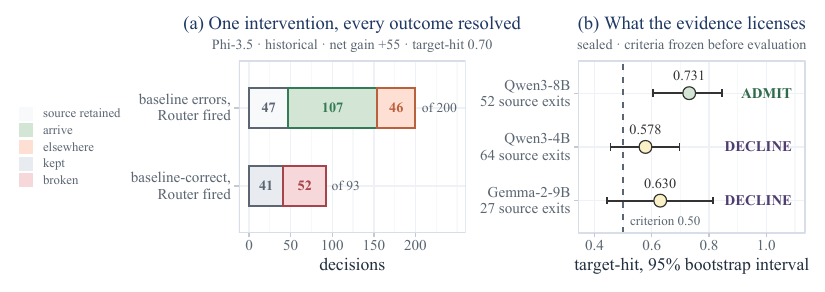}
    \vspace{-0.25in}
    \caption{\textbf{Correction is not repair.} \textbf{(a)} Outcome-resolved steering on Phi-3.5-mini. Of 200 routed errors, 107 reach the target while 46 shift into alternative errors; concurrently, 52 of 93 exposed correct decisions break, despite a net $+55$ gain. \textbf{(b)} Target-hit rates ($95\%$ bootstrap CIs) against the frozen $0.50$ criterion. Qwen3-8B clears the gate; Qwen3-4B and Gemma-2-9B yield favorable point estimates whose confidence intervals cross the threshold, warranting formal \textsc{DECLINE} verdicts due to evidentiary shortfall rather than uncorrectability.}
    \vspace{-1em}
    \label{fig:thesis}
\end{figure}

While \S~\ref{sec:results} confirms genuine, direction-specific behavioral steering, we assess whether such shifts warrant a formal claim of repair. In multiclass action spaces, transitions do not collapse into binary shifts. Hence, licensing a repair requires clearing three independent, non-trivial hurdles beyond raw steerability: (1)~\textit{destination correctness} (reaching gold target), (2)~\textit{preservation} (sparing baseline-correct decisions), (3)~\textit{evidential sufficiency} (requiring the confidence interval, not only the point estimate, to clear the prospective threshold). Crucially, each criterion can fail independently even when prior conditions are met.

\subsection{Decoupling Correction from Destination Correctness}
\label{subsec:destination}

\begin{figure}[t]
    \centering
    \vspace{-1em}
    \includegraphics[width=0.88\linewidth]{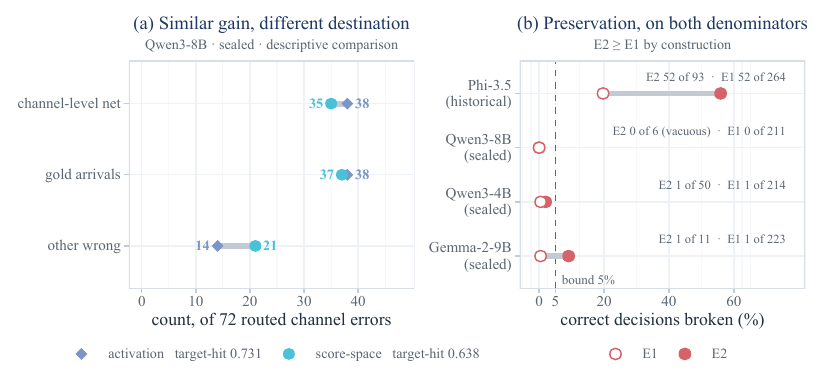}
    \vspace{-0.3in}
    \caption{\textbf{Destination correctness and preservation are separate properties.} \textbf{(a)} On Qwen3-8B, activation steering and logit shifting yield similar target arrivals but differ descriptively in lateral error spillage (14 vs.\ 21 exits; $\mathrm{TH}_c = 0.731$ vs.\ $0.638$), without surviving multiplicity correction (McNemar $p = 1.0$ and $p = 0.189$). \textbf{(b)} Collateral damage across denominators vs.\ a $5\%$ tolerance: Phi-3.5-mini incurs $52/93$ exposed ($\mathrm{E2}$) and $52/264$ overall ($\mathrm{E1}$) breaks; Qwen3-8B incurs $0/211$ overall (6 exposed); Qwen3-4B incurs $1/50$ exposed; and Gemma-2-9B incurs $1/11$ exposed ($\mathrm{E2} = 9.1\% > 5\%$). By construction, $\mathrm{E2} \ge \mathrm{E1}$. (See Figure~\ref{fig:thesis}b for licensing intervals).}
    \vspace{-1.5em}
    \label{fig:destination}
\end{figure}

\textbf{Auditing Multiclass Exits and Lateral Error Mass Redistribution.}
Row-level destination audits reveal substantial lateral probability leakage into non-target error classes across models. On Phi-3.5-mini, 153 of 200 routed errors successfully vacate the source state ($X_c = 153$); however, while 107 reach the gold target action, 46 spill into alternative error classes, yielding a target-hit rate of $\mathrm{TH}_c = 0.699$ (Fig.~\ref{fig:thesis}a). On Qwen3-8B, internal activation steering and external logit shifting achieve near-identical top-line gains ($+38$ vs.\ $+35$ net gain; 38 vs.\ 37 gold arrivals; Fig.~\ref{fig:destination}a) but diverge in destination topology: internal steering misdirects only 14 of 52 exits into alternative errors ($\mathrm{TH}_c = 0.731$), whereas logit shifting diverts 21 of 58 exits into off-target failures ($\mathrm{TH}_c = 0.638$).

\textbf{Vacating Error States Does Not Imply Target Convergence.}
Exiting an error state does not guarantee ground-truth convergence in multiclass settings. On Phi-3.5-mini, nearly a third of exits (46 of 153) drift into alternative errors rather than gold target, i.e., a defect entirely concealed by aggregate metrics. On Qwen3-8B, matching headline gains obscure diverging destination profiles (a descriptive contrast asserting no formal ordering). By auditing target-hit rate, \textsc{SAKIKO} formally decouples source departure from verified target arrival, resolving what standard metrics conflate.

\subsection{Decoupling Destination Correctness from Preservation}
\label{subsec:preservation}

\textbf{Collateral Breakdown Profiles and Denominator-Dependent Degradation.}
Audits of baseline-correct instances expose severe collateral damage unmitigated by router gating. On Phi-3.5-mini, steering corrupts 52 of 93 exposed correct decisions (Fig.~\ref{fig:thesis}a), yielding acute ($\mathrm{E2} = 55.9\%$) and population ($\mathrm{E1} = 19.7\%$) losses far exceeding tolerance ($\epsilon_{\mathrm{tol}} = 0.05$). Router confidence fails to separate broken from intact instances (Cliff's $\delta = -0.094$, $p = 0.44$), leaving break rates $>50\%$ even at high thresholds ($\tau_c = 0.9$). Moreover, safety conclusions hinge critically on denominator selection: on Qwen3-4B, a benign population loss ($\mathrm{E1} = 1.3\%$) conceals acute localized degradation ($\mathrm{E2} = 5.7\%$), whereas on Qwen3-8B, zero breaks produce an uninformative exposed bound ($39.3\%$) owing to sparse router exposure ($\vert{}C_{\mathrm{exp}}\vert{} = 6$).

\textbf{The Orthogonality of Target Steering and Behavioral Invariance.}
Successful steering on error instances provides no inherent safeguard against corrupting clean inputs. Latent overlap between under- and fully specified prompts causes inference-time routers to misallocate steering vectors, and raising confidence thresholds fails to mitigate break rates among exposed decisions. Because breaks fall entirely within exposed instances, population loss $\mathrm{E1}$ understates localized risk by $\vert{}C\vert{} / \vert{}C_{\mathrm{exp}}\vert{}$, necessitating reporting both $\mathrm{E1}$ and exposed loss $\mathrm{E2}$ alongside raw counts. However, because preregistered protocols formally fixed preservation on $\mathrm{E1}$ (\texttt{clean\_collateral\_rate} on Qwen3-8B, with exposure-conditional metrics added only in subsequent runs), we report $\mathrm{E2}$ without retrofitting frozen rules. Consequently, the resulting repair license guarantees preservation strictly at the population level, not conditional on router exposure.

\subsection{Prospective Licensing and Evidential Sufficiency}
\label{sec:licensing}

\textbf{Empirical Adjudication Under Pre-Registered Statistical Gating.}
\textsc{SAKIKO}'s frozen ten-condition gate dissociates nominal point estimates from statistically robust repairs. Qwen3-8B clears all criteria for an \textsc{ADMIT}, maintaining confidence bounds strictly above required thresholds ($\mathrm{TH}_c = 0.731$, CI $[0.604, 0.846]$; $\mathrm{TG}_c = 0.276$, CI $[0.115, 0.425]$) with zero breaks ($B = 0$). In contrast, Qwen3-4B and Gemma-2-9B receive \textsc{DECLINE} verdicts: despite positive point metrics, their bootstrap intervals cross sub-threshold bounds on target-hit rate and span zero on target gain ($\mathrm{TG}_c$ CIs $[-0.048, 0.202]$ and $[-0.031, 0.177]$, respectively). On Gemma-2-9B, wide intervals breach Conditions 3 and 6, while an acute exposed loss of $\mathrm{E2} = 9.1\%$ ($1/11$ breaks) underscores that a nominal preservation pass does not guarantee deployment safety.

\textbf{Bounding Optimism via Finite-Sample Statistical Power.}
Finite-sample point estimates cannot prove claims of mechanistic repair. Due to limited statistical power, resolving a modest hit rate of $0.58$ against a $0.50$ null requires $245$ exits (versus ${\sim}30$ for $0.73$), reflecting inadequate precision rather than guaranteed admission under larger samples. Both model declines persist even if hit-rate gates are relaxed, as target-gain intervals cross zero and exposed collateral damage remains uncontained. By mandating that confidence intervals, not isolated point estimates, satisfy preregistered safety criteria, \textsc{SAKIKO} ensures small-sample optimism is never mistaken for validated repair.

\begin{figure}[t]
    \centering
    \vspace{-1em}
    \includegraphics[width=0.88\linewidth]{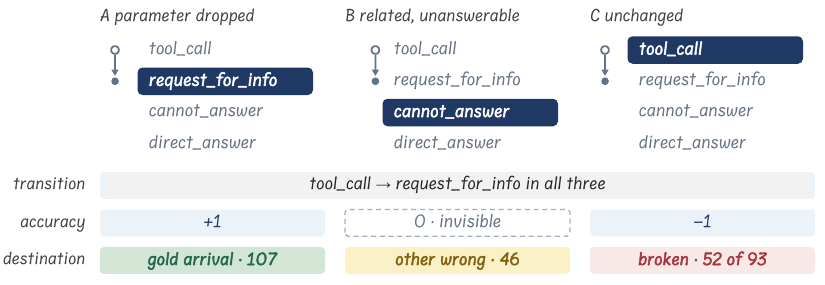}
    \vspace{-.2in}
    \caption{One question under three input variants (panels A--C). Open circles and solid dots indicate pre- and post-intervention decisions, showing an identical shift across panels (\texttt{tool\_call} $\rightarrow$ \texttt{request\_for\_info}). Filled labels denote the ground-truth target; in panel B, the post-intervention dot lands above the target, capturing an exit without target arrival. The bottom bands contrast readout resolutions: transition is unsegmented; accuracy captures A and C but misses B (consistently incorrect); destination resolves all three distinct outcomes. Visualizations are illustrative; counts represent frozen outcome-class totals rather than specific transition frequencies.}
    \vspace{-1em}
    \label{fig:casestudy}
\end{figure}

\subsection{One Transition, Three Verdicts}
\label{subsec:casestudy}

\textbf{Uniform Output Transitions Across Heterogeneous Task Demands.}
On When2Call, an identical intervention shift, i.e., from invoking \texttt{search\_flights} to requesting information, yields three distinct outcome classes depending on context, illustrated for Phi-3.5-mini in Fig.~\ref{fig:casestudy}. In Case~A (\textit{missing origin}), the model suppresses a hallucinated departure city to ask for clarification, achieving a target repair (\textsc{GOLD ARRIVAL}, 107 instances). In Case~B (\textit{unserviceable status inquiry}), it leaves the tool call but requests an unneeded flight number instead of refusing, shifting laterally into a new error (\textsc{OTHER WRONG}, 46 instances). In Case~C (\textit{fully specified query}), an already-correct tool invocation is needlessly routed and broken into redundant clarification (\textsc{BROKEN}, 52 of 93 exposed instances). Standard accuracy tallies these identical shifts as $+1$, $\pm 0$, and $-1$, collapsing qualitatively divergent behaviors into an aggregate net gain of $+55$ ($107 - 52$).

\textbf{Representational Opacity of Behavioral Uniformity.}
Uniform action-level shifts mask fundamentally divergent epistemic outcomes, e.g., spanning true repair, lateral error drift, and collateral corruption. Standard aggregate accuracy cannot separate these phenomena: it ignores lateral errors (Case~B) while permitting true repairs (Case~A) and broken decisions (Case~C) to neutralize each other. Because top-line metrics conceal lateral probability redistribution and collateral damage, validating representation repair requires row-level, destination-resolved auditing across entire $K$-way action space.

\section{Conclusion}
\label{sec:conclusion}
In autonomous agents, pre-execution action errors represent silent failures rooted in internal representations. We introduced \textbf{SAKIKO}, a destination-resolved framework that formally tests when representation interventions constitute genuine repair. While channel-keyed steering induces direction-specific corrections across five architectures, conventional net gains conceal crucial failure modes: transitions frequently misroute into lateral errors, degrade baseline-correct decisions, and fail formal confidence bounds, i.e., declining two of three sealed models on destination intervals alone despite favorable point estimates. By pairing outcome-resolved tracking with preregistered statistical gating, SAKIKO separates certified interventions from cosmetic behavioral drift. Crucially, this license certifies only categorical action-mode selection rather than downstream execution validity or exposed decision safety (App.~\ref{sec:limitations}), establishing that rigorous behavioral repair requires auditing destinations, collateral damage, and uncertainty alongside top-line gains.

\section*{Reproducibility Statement}
Appendices~\ref{app:models} and~\ref{app:protocol};
Appendix~\ref{app:repro} gives the artifact locations, the exact invocation, the
determinism controls and the Git LFS requirement. The sealed evaluations were each executed once
under a preregistered gate, and the one-shot ledger is recorded in
Appendix~\ref{app:provenance}. The code is available in~\url{https://github.com/ruizheliUOA/mechanistic-tool-use-llm}.

\bibliography{iclr2027_conference}
\bibliographystyle{iclr2027_conference}

\appendix
\section{Limitations}
\label{sec:limitations}

We explicitly delineate the boundaries of our empirical findings across five primary areas:

\textbf{Benchmark Scope and External Transfer.} All formal confirmatory evaluations are established exclusively on When2Call \citep{ross2025when2call}. While MetaTool provides historical evidence in a binary setting where destination analysis is structurally inapplicable, ACEBench failed pre-intervention channel-support criteria and was retired prior to steering experiments. Furthermore, because ground-truth labels in When2Call are synthetically generated by its creators without post-hoc re-annotation, unmodeled systematic labeling biases could shift destination distributions without triggering gate violations. Cross-benchmark, multi-modal, or real-world policy transfer remains unestablished.

\textbf{Specificity and Scope of Licensed Claims.} Our positive repair claim rests upon a single admitted configuration: Qwen3-8B on the \texttt{cannot\_answer}~$\rightarrow$~\texttt{tool\_call} channel under gradient activation steering. This outcome does not support general claims regarding the model family, other error transitions, or activation steering broadly. Furthermore, activation interventions are not established as superior to score-space comparators: paired tests after multiplicity correction remain non-significant (exact McNemar $p = 1.0$ and $p = 0.189$), and performance rankings between the two alternate across architectures. Existing steering methods (e.g., ASA, CAST) were not re-implemented as empirical baselines because our evaluation protocol was strictly locked prior to sealed execution.

\textbf{Exposure-Conditional Preservation Deficits.} Although our pre-registered gate bounded population-level collateral damage ($E1 \le 0.05$), no setting in this study certifies exposure-conditional preservation ($E2$). On Qwen3-8B, zero breaks were observed over only six router-exposed correct decisions, yielding an uninformative one-sided $95\%$ upper bound of $0.393$ (whereas certifying a $5\%$ bound requires at least 59 exposed correct instances). Similarly, Gemma-2-9B corrupted $1$ of $11$ exposed correct decisions ($E2 = 0.091$) despite passing the overall population threshold. Demonstrating localized safety on touched clean decisions remains open across all evaluated models.

\textbf{Sensitivity to Sample Support and Finite-Sample Uncertainty.} In sealed evaluations, formal declines for Qwen3-4B and Gemma-2-9B were driven entirely by bootstrap interval conditions rather than point-estimate failures. The outcome-resolved apparatus provides the necessary statistical resolution to compute these intervals, confirming that small-sample optimism cannot substitute for verified repair. Similarly, the pre-intervention halt of Qwen3.5-9B (at 29 reference rows versus a 30-instance threshold) was split-sensitive, clearing eligibility in $56.7\%$ of retrospective split reallocations rather than reflecting intrinsic uncorrectability.

\textbf{Mechanistic Scope and Procedural Transparency.} While we confirm site- and direction-specific causal steerability, our framework audits the empirical evidence pipeline rather than isolating causal circuits or tracing detailed propagation dynamics through model components. Finally, to ensure complete provenance, we disclose that three procedural gate adjustments, one mechanical restart on Qwen3-4B, and historical variations in random null sizes (1 to 10 directions on early Qwen2.5-7B runs versus 59 in sealed settings) occurred. All events were cryptographically recorded with pre-access hashes prior to endpoint unblinding without altering any frozen verdict.

\section{Problem Formalisation and Notation}
\label{app:notation}

This appendix fixes the populations and denominators used throughout the paper.
Ambiguity about which population a rate is computed over has been the single
largest source of misreading in this line of work, so every quantity below is
defined against a named set rather than against ``the evaluation set''.

\subsection{Actions, channels and populations}

Let $\mathcal{Y}$ be the set of pre-execution action modes available to the model.
On the primary benchmark $|\mathcal{Y}| = 4$: calling a tool, requesting missing
information, answering directly, and declining. For an input $x$ we write
$y^\star(x) \in \mathcal{Y}$ for the reference action and $\hat{y}_0(x)$ for the
action the unmodified model produces. An input is a \emph{baseline error} when
$\hat{y}_0 \neq y^\star$ and \emph{baseline-correct} when $\hat{y}_0 = y^\star$.

A \textbf{directional error channel} is the ordered pair
\begin{equation}
    c \;=\; \bigl(g \rightarrow s\bigr), \qquad g = y^\star,\quad s = \hat{y}_0,
    \quad g \neq s ,
    \label{eq:app-channel}
\end{equation}
where $g$ is the required action and $s$ the source action actually produced.
Channels are properties of a model on a task: the transition carrying the most
error mass differs between models, so the framework discovers them
(Appendix~\ref{app:protocol}) rather than assuming a fixed set.

Four populations matter, and they are not interchangeable (Table~\ref{tab:notation} collects every symbol with its denominator):
\begin{align}
    E_c          &= \{x : \hat{y}_0(x) = s,\; y^\star(x) = g\}
                    && \text{channel errors, } N_c = |E_c| \label{eq:app-pop-e}\\
    E_c^{\mathrm{rt}} &= \{x \in E_c : \mathrm{fire}(x)\}
                    && \text{routed channel errors} \label{eq:app-pop-rt}\\
    C            &= \{x : \hat{y}_0(x) = y^\star(x)\}
                    && \text{all baseline-correct decisions} \label{eq:app-pop-c}\\
    C_{\mathrm{exp}} &= \{x \in C : \mathrm{fire}(x)\}
                    && \text{exposed (at-risk) correct decisions} \label{eq:app-pop-exp}
\end{align}
with $E_c^{\mathrm{rt}} \subseteq E_c$ and $C_{\mathrm{exp}} \subseteq C$. The
distinction between \eqref{eq:app-pop-e} and \eqref{eq:app-pop-rt}, and between
\eqref{eq:app-pop-c} and \eqref{eq:app-pop-exp}, is what makes the rates below
well defined. On Qwen3-8B, for instance, $N_c = 87$ while
$|E_c^{\mathrm{rt}}| = 72$; a count reported ``of 72'' is not the same statement as
the same count reported ``of 87''.

\subsection{Post-intervention outcome classes}

The intervention produces a second action $\hat{y}_1$. Every decision in the two
adjudicated cohorts --- the errors of channel $c$ and the baseline-correct
population --- falls into exactly one of five classes, three on the error
population and two on the correct one:
\begin{equation}
    \underbrace{N_c = S_c + A_c + O_c}_{\text{baseline errors}},
    \qquad
    \underbrace{|C_{\mathrm{exp}}| = R + B}_{\text{exposed correct}},
    \label{eq:app-partition}
\end{equation}
where $S_c$ retain the source ($\hat{y}_1 = s$), $A_c$ arrive at the required
action ($\hat{y}_1 = g$), $O_c$ exit to a third action
($\hat{y}_1 \notin \{g, s\}$), $R$ remain correct and $B$ are broken. Decisions the
Router declines are returned unchanged. An unrouted channel error therefore
retains its source and counts in $S_c$, which is defined over all of $N_c$;
an unrouted correct decision is by definition outside $C_{\mathrm{exp}}$ and
counts in neither $R$ nor $B$, both of which are defined over the exposed cohort
alone. We write $X_c = A_c + O_c$ for the \emph{source exits}.

\begin{table}[t]
    \centering
    \small
    
    \caption{The five mutually exclusive post-intervention outcome classes. Aggregate net gain ($G = \sum_c A_c - B$) leaves exits to other incorrect actions ($O_c$) unobserved and allows target arrivals ($A_c$) and collateral breaks ($B$) to artificially cancel.}
    \label{tab:outcomes}
    \resizebox{0.7\textwidth}{!}{
    \begin{tabular}{lll}
        \toprule
        Population & Outcome Class & Condition \\
        \midrule
        \multirow{3}{*}{\shortstack[l]{Baseline Channel Errors\\($N_c$, where $\hat{y}_0 = s$)}} 
        & \textsc{SOURCE RETAINED} ($S_c$) & $\hat{y}_1 = s$ \\
        & \textsc{GOLD ARRIVAL} ($A_c$)     & $\hat{y}_1 = g$ \\
        & \textsc{OTHER WRONG} ($O_c$)      & $\hat{y}_1 \notin \{g, s\}$ \\
        \midrule
        \multirow{2}{*}{\shortstack[l]{Exposed Baseline Correct\\($C_{\mathrm{exp}}$, where $\hat{y}_0 = y^\star$)}} 
        & \textsc{CORRECT RETAINED}        & $\hat{y}_1 = y^\star$ \\
        & \textsc{BROKEN} ($B$)            & $\hat{y}_1 \neq y^\star$ \\
        \bottomrule
    \end{tabular}}
    \vspace{-1em}
\end{table}

\subsection{Endpoints}

\begin{align}
    G_{\mathrm{whole}} &= \text{Fixed} - B,
    \quad \text{Fixed} = \bigl|\{x : \hat{y}_0 \neq y^\star,\ \hat{y}_1 = y^\star\}\bigr|
    &&\text{whole-population net}
    \label{eq:app-net}\\[3pt]
    G_{\mathrm{channel}} &= \textstyle\sum_c A_c - B
    &&\text{channel-level net}
    \label{eq:app-netc}\\[3pt]
    \mathrm{TH}_c &= \frac{A_c}{A_c + O_c} = \frac{A_c}{X_c}
    &&\text{target-hit; undefined at } X_c = 0
    \label{eq:app-th}\\[3pt]
    \mathrm{TG}_c &= \frac{A_c - O_c}{N_c}
    &&\text{target gain, over \emph{all} channel errors}
    \label{eq:app-tg}\\[3pt]
    \mathrm{E1} &= \frac{B}{|C|},
    \qquad
    \mathrm{E2} = \frac{B}{|C_{\mathrm{exp}}|}
    &&\text{collateral, two denominators}
    \label{eq:app-collateral}
\end{align}

\paragraph{The two net conventions are distinct and are never substituted.}
$\mathrm{Fixed}$ counts every baseline error in the evaluation population that
becomes correct; $\sum_c A_c$ counts only arrivals inside the adjudicated
channels. The first exceeds the second whenever the intervention corrects an
error the channel does not key, so $G_{\mathrm{whole}} \geq G_{\mathrm{channel}}$,
with equality only when no such error is corrected. On Qwen3-8B the two are $+40$
and $+38$ (Table~\ref{tab:q8b-flow}). Every net quantity in this paper is
labelled \textit{whole} or \textit{channel-level} and the two are never
interchanged.

Three consequences follow directly and are used throughout the paper.

\paragraph{Net compresses five classes into two.} Only $A_c$ and $B$ appear in
\eqref{eq:app-net} and \eqref{eq:app-netc}. $O_c$ does not, because a move from one wrong action to
another leaves accuracy unchanged. A given $G$ therefore does not determine
$\mathrm{TH}_c$ or the collateral: for any $G$ and any rational $t \in (0,1]$ whose
denominator divides $A_c$, setting $O_c = A_c(1-t)/t$ gives target-hit exactly $t$
at unchanged $G$, and every such profile is admissible whenever the population is
large enough to contain it. The mapping from the five classes to $G$ is
many-to-one; that, and not an absence of any constraint, is the formal content of
the paper's central claim.

\paragraph{The two collateral denominators are ordered.} Breaks occur only within
$C_{\mathrm{exp}}$, and $C_{\mathrm{exp}} \subseteq C$, so
\begin{equation}
    \mathrm{E2} = \frac{B}{|C_{\mathrm{exp}}|} \;\geq\; \frac{B}{|C|} = \mathrm{E1},
    \label{eq:app-e2ge1}
\end{equation}
with equality when $B = 0$ or when every correct decision is exposed, and strict
inequality otherwise. Reporting E1 alone
understates the risk to a touched decision by the factor
$|C| / |C_{\mathrm{exp}}|$. The frozen licensing gate uses E1, the permissive
denominator; we report both throughout, with their counts.

\paragraph{Target-hit is degenerate in a binary action space.} If
$|\mathcal{Y}| = 2$ then $O_c = 0$ by construction and $\mathrm{TH}_c \equiv 1$
whenever $X_c > 0$. The destination question is therefore not posable on a binary
benchmark, which is why the destination-resolved results in this paper come from a
multiclass action space (Appendix~\ref{app:data}).

\begin{table}[t]
    \centering
    \small
    \setlength{\tabcolsep}{6pt}
    \renewcommand{\arraystretch}{1.18}
    \caption{Notation and endpoint definitions. The \emph{population} column is
    the denominator; quoting a rate without it is ambiguous. Populations are
    defined in \eqref{eq:app-pop-e}--\eqref{eq:app-pop-exp}.}
    \label{tab:notation}
    \begin{tabular}{@{}llll@{}}
        \toprule
        Symbol & Quantity & Population (denominator) & Defined in \\
        \midrule
        \addlinespace[1pt]
        $c = (g \rightarrow s)$ & directional error channel & --- & \eqref{eq:app-channel} \\
        $N_c$ & channel errors & --- & \eqref{eq:app-pop-e} \\
        $E_c^{\mathrm{rt}}$ & routed channel errors & --- & \eqref{eq:app-pop-rt} \\
        $C$ & baseline-correct decisions & --- & \eqref{eq:app-pop-c} \\
        $C_{\mathrm{exp}}$ & exposed correct decisions & --- & \eqref{eq:app-pop-exp} \\
        \addlinespace[4pt]
        $S_c$ & source retained & $N_c$ & \eqref{eq:app-partition} \\
        $A_c$ & gold arrival & $N_c$ & \eqref{eq:app-partition} \\
        $O_c$ & other wrong & $N_c$ & \eqref{eq:app-partition} \\
        $X_c$ & source exits, $A_c + O_c$ & $N_c$ & \eqref{eq:app-partition} \\
        $R$ & correct retained & $|C_{\mathrm{exp}}|$ & \eqref{eq:app-partition} \\
        $B$ & broken & $|C_{\mathrm{exp}}|$ & \eqref{eq:app-partition} \\
        \addlinespace[4pt]
        $G_{\mathrm{whole}}$ & whole-population net & evaluation set & \eqref{eq:app-net} \\
        $G_{\mathrm{channel}}$ & channel-level net & adjudicated channels & \eqref{eq:app-netc} \\
        $\mathrm{TH}_c$ & target-hit & source exits $X_c$ & \eqref{eq:app-th} \\
        $\mathrm{TG}_c$ & target gain & channel errors $N_c$ & \eqref{eq:app-tg} \\
        $\mathrm{E1}$ & collateral, population & all correct $|C|$ & \eqref{eq:app-collateral} \\
        $\mathrm{E2}$ & collateral, exposure-conditional & exposed $|C_{\mathrm{exp}}|$ & \eqref{eq:app-collateral} \\
        \bottomrule
    \end{tabular}
\end{table}

%
%
\section{Extended Related Work}\label{app:related_work}

We draw on three lines of work: tool-augmented agents, interpretability of agent decisions, and representation-level intervention. In this section we survey each and delineate the structural distinctions that characterize \textsc{SAKIKO}.

\subsection{Tool-Augmented Agents and the Epistemics of Action Selection}
\label{subsec:rw_tool_learning}

The integration of external tools into LLMs has significantly expanded their functional scope from static text generators to autonomous problem solvers~\citep{wei2022emergent,qin2024tool}. Early and influential paradigms, such as Toolformer~\citep{schick2023toolformer}, Gorilla~\citep{patil2024gorilla}, ToolLLM~\citep{qin2024toolllm}, and ToolAlpaca~\citep{tang2023toolalpaca}, focused predominantly on the syntactic and parametric execution of tool calls, i.e., improving how models format API queries, satisfy parameter schemas, and adapt to tool documentation via fine-tuning or in-context demonstration.

However, before an agent can parameterize or execute a specific tool, it must resolve the fundamentally upstream decision of \textit{whether} external action is warranted at all. A recent conceptual framework proposed by~\citet{wang2026positionagentinvokeexternal} formalizes this criterion, positing that agents should invoke external tools only when epistemically necessary, i.e., when task uncertainty cannot be resolved through internal reasoning alone. Deviations from this criterion give rise to failure regimes such as tool bypass (unwarranted direct answering despite missing evidence) or over-delegation (unnecessary external queries when internal knowledge suffices). While~\citet{wang2026positionagentinvokeexternal} conceptually formalize the normative decision boundary between internal reasoning and external interaction, they do not study the underlying internal representation geometry nor provide intervention mechanisms to correct miscalibrated tool decisions. \textsc{SAKIKO} provides an empirical and representation-level operationalization of this boundary, identifying and intervening on the directional channels that mediate pre-execution tool decisions.

A parallel line treats the same decision as a policy to be trained rather than a
representation to be edited. \citet{zhou2026trust} optimise tool-calling decisions
by uncertainty-aligned reinforcement learning over the same four next-action
choices, and \citet{suri2026clarification} select clarification questions from a
structured uncertainty estimate, reporting large When2Call gains from
uncertainty-weighted training. \citet{liu2026agentabstain} evaluate calibrated
restraint directly, pairing should-act with should-abstain variants of the same
task. These are alternatives to post-hoc intervention rather than comparators for
it: they change the policy, where \textsc{SAKIKO} leaves the weights frozen and
asks what evidence an edit to a frozen model's activations can support. None of
them resolves where a changed decision lands among the remaining actions, which
is the accounting this paper adds.

\subsection{Internal Observability and Detection of Tool-Use Failures}
\label{subsec:rw_observability}

Paralleling the rise of agent frameworks, recent work in mechanistic interpretability has begun investigating how internal representations reflect an agent's operational state prior to execution~\citep{yan2026spurious}. Traditional reliability taxonomies primarily address factual hallucinations in ungrounded text generation~\citep{ji2023survey,li2026attributing}. In contrast, agentic settings introduce structured failures in action selection.

To monitor these failures,~\citet{healy2026internalrepresentationsindicatorshallucinations} investigate internal representations during tool-call generation, demonstrating that latent embeddings in transformer layers linearly encode tool-selection and parameter hallucinations. By training lightweight binary classifiers on contextualized hidden states, they show that impending tool errors can be flagged in real time before execution. Concurrently,~\citet{tatsat2026blackboxinterpretabilityagentic} construct an interpretability framework that couples sparse autoencoders (SAEs) with linear probes to monitor agent states before action. Their architecture deploys a binary tool-need probe to predict tool invocation and a ternary tool-risk probe to estimate operational consequence, demonstrating that decision signals localize to specific sparse features and late transformer layers.

While these approaches substantiate that an agent's pre-execution intent and validity are decodable from internal states, they remain strictly diagnostic: they focus on passive monitoring, risk scoring, and post-hoc feature localization. Furthermore, their formulation of error detection is typically collapsed into a binary classification (e.g., correct vs.\ hallucinated, or tool needed vs.\ not needed). \citet{zhao2026calibration} are the exception, decomposing multi-turn failures over a four-class action space into action-class miscalibration and execution error, but they diagnose the miscalibration without intervening on the states that produce it; \citet{wu2026call} do drive a controller from such readouts, but score the result by task performance rather than by where the redirected decision lands. \citet{shi2026overcalling} go further and intervene, recovering a sparse-autoencoder
feature basis for the call/no-call decision on When2Call, reducing it to a signed
activation margin, and testing an activation-independent CALL offset causally by a
closed-form counter-bias shift along the decoder directions. That study and this
one share a benchmark and an intervention family and ask different questions: it
diagnoses and cancels a scalar bias on a binary margin, where vacating the source
leaves one destination, while the questions here --- which of three or more
remaining actions a corrected decision reaches, and what the correction costs on
decisions that were already right --- are not posable in that formulation.
Outside tool use, \citet{jiang2026behavioural} make the complementary point that a
model can pass behavioural safety evaluation while remaining vulnerable to bounded
latent perturbation, which is the same dissociation between an output-level
readout and an intervention-level one that motivates our destination accounting.
\textsc{SAKIKO} departs from passive observability by developing active, targeted interventions. More fundamentally, \textsc{SAKIKO} recognizes that pre-execution tool decisions operate in a multi-way action space, where diagnostic flags alone cannot resolve which directional channel caused the failure or guide the model toward the correct target action.

\subsection{Representation Engineering and Activation Steering in Agents}
\label{subsec:rw_steering}

Representation engineering and activation steering have emerged as efficient, backbone-training-free techniques to manipulate model behaviour by perturbing intermediate residual states~\citep{zou2025representationengineeringtopdownapproach}. In the context of alignment and behavioural control, Conditional Activation Steering (CAST)~\citep{lee2025programming} demonstrates that refusal mechanisms can be gated conditionally using similarity metrics derived from prompt activations.

Recently, representation steering has been extended directly to agent tool selection. \citet{wu2026tool} demonstrate that tool identity is linearly readable and steerable within the residual stream across a wide spectrum of open-weight models. By injecting the mean-difference activation vector between two candidate tools, they show that an agent's discrete tool choice can be flipped with high accuracy on single-turn menus, with the autoregressively generated JSON arguments updating to match the schema of the new tool. Similarly, the Activation Steering Adapter (ASA)~\citep{wang2026asa} applies router-conditioned steering vectors at mid-layers to counter the lazy agent problem, steering models out of inert states when tool invocation is warranted.

Despite these empirical successes, the steering methodologies we survey operate under a crucial simplification: they evaluate steering as either a binary transition (e.g., refusal vs.\ compliance in CAST, or tool vs.\ no-tool in ASA) or as an isolated pairwise swap between two designated tools~\citep{wu2026tool}. In such constrained environments, vacating an incorrect state trivially coincides with arriving at the alternative candidate. As we demonstrate, this assumption does not hold in the multiclass settings we evaluate.

\subsection{Contrasting SAKIKO: From Behavioural Movement to Adjudicated Repair}
\label{subsec:rw_contrast}

\begin{table}[t]
\centering
\footnotesize
\setlength{\tabcolsep}{3pt}
\caption{\textbf{Comparison of \textsc{SAKIKO} with the agent interpretability and steering frameworks we survey.} Unlike the passive diagnostic tools and binary or pairwise steering methods listed here, \textsc{SAKIKO} addresses multiclass action spaces, tracks post-intervention destination distributions, verifies baseline preservation, and formally adjudicates repair claims under sample uncertainty. \emph{Dest.} is destination verification and \emph{Lic.} is evidential licensing. The four-class row is diagnostic only: it decomposes failures without intervening, so no destination distribution is produced to verify.}
\label{tab:related_work_comparison}
\begin{tabular}{@{}p{0.28\linewidth}p{0.26\linewidth}p{0.24\linewidth}cc@{}}
\toprule
\textbf{Methodology} & \textbf{Intervention Paradigm} & \textbf{Action Space} & \textbf{Dest.} & \textbf{Lic.} \\
\midrule
ToA Position~\citep{wang2026positionagentinvokeexternal} & Conceptual framework & Epistemic boundary & \texttimes & \texttimes \\
Tool Hallucination Probes~\citep{healy2026internalrepresentationsindicatorshallucinations} & Passive linear probing & Binary (error / valid) & \texttimes & \texttimes \\
Beyond the Black Box~\citep{tatsat2026blackboxinterpretabilityagentic} & Passive SAE $+$ probing & Binary / ternary risk & \texttimes & \texttimes \\
Action-Class Diagnostic~\citep{zhao2026calibration} & Diagnostic decomposition (no intervention) & Four-class & \texttimes & \texttimes \\
CAST~\citep{lee2025programming} & Conditional steering & Binary (refuse / comply) & \texttimes & \texttimes \\
Tool Steering Probes~\citep{wu2026tool} & Additive steering & Pairwise (tool A $\to$ B) & \texttimes & \texttimes \\
ASA~\citep{wang2026asa} & Router-conditioned steering & Binary (call / bypass) & \texttimes & \texttimes \\
To Call or Not to Call~\citep{wu2026call} & Hidden-state readout $\to$ controller & Binary (call / bypass) & \texttimes & \texttimes \\
Over-Calling Bias~\citep{shi2026overcalling} & SAE margin $+$ counter-bias steering & Binary (call / no-call) & \texttimes & \texttimes \\
\midrule
\textbf{\textsc{SAKIKO} (ours)} & \textbf{Channel-keyed steering} & \textbf{$K$-way action space} & \checkmark & \checkmark \\
\bottomrule
\end{tabular}
\end{table}

Table~\ref{tab:related_work_comparison} highlights the structural position of \textsc{SAKIKO} relative to the works we survey. While this research establishes that tool intent is linearly decodable~\citep{healy2026internalrepresentationsindicatorshallucinations,tatsat2026blackboxinterpretabilityagentic} and steerable~\citep{wang2026asa,wu2026tool}, these studies evaluate intervention success through aggregate metric shifts or binary state exits. \textsc{SAKIKO} addresses three issues that the surveyed evaluations do not:

\begin{enumerate}[leftmargin=*, wide=0pt, itemsep=2pt, topsep=1pt]
    \item \textbf{Multiclass Destination Divergence ($K$-way Outcomes):} Real-world pre-execution agent decisions are not binary, and they span invoking specific tools, requesting user clarification, answering directly, or declining inappropriate prompts. In a $K$-way action space, perturbing an internal activation away from an erroneous state does not guarantee arrival at the ground-truth target. An intervention can induce substantial behavioural movement while simply redirecting the model to an alternative, equally incorrect action (e.g., misdirecting a tool-bypass error into a refusal). \textsc{SAKIKO} explicitly tracks the full destination composition rather than treating state-departure as repair.

    \item \textbf{Decoupling Net Gain from Baseline Preservation:} Prior steering works report aggregate improvements (e.g., overall tool-calling accuracy) across evaluation sets. However, as our experiments reveal (e.g., on Phi-3.5), an intervention can produce a net gain of $+55$ while concurrently corrupting $52$ of the $93$ already-correct decisions on which its detector fired. By auditing both destination hit-rates and collateral damage on the baseline-correct inputs the intervention actually touches, \textsc{SAKIKO} ensures that repair does not come at the expense of unmonitored capability degradation.

    \item \textbf{Evidence-Qualified Licensing under Uncertainty:} In the works we survey, favourable point estimates are generally reported without an accompanying uncertainty criterion. However, under finite evaluation budgets and high variance, point gains can be statistically illusory. \textsc{SAKIKO} introduces a structured adjudication protocol that combines destination verification with pre-registered uncertainty criteria, formally declining repair claims when empirical sample support is insufficient.
\end{enumerate}

\paragraph{Adjacent evaluation settings.} Three further lines bound what the
present evaluation covers rather than compete with it.
\citet{agarwal2025toolrm} train outcome reward models for tool calling and
introduce a reward benchmark for them, which scores whether a call was good rather
than whether an action mode was the right one; linking pre-execution repair to
outcome quality is the natural next step and one this paper does not take.
\citet{li2026toolprmbench} supply step-level process labels over trajectories,
a richer setting than the single pre-execution decision audited here.
\citet{luo2026multilingual} show that tool-calling failures shift systematically
across languages, and \citet{laskar2026texttovoice} that they shift again when the
same benchmarks are rendered as speech --- including When2Call, which carries every
formal outcome in this paper. Both mark a boundary on the generality of our
results: nothing here establishes that the routing and collateral findings persist
beyond English text prompts.

Therefore, rather than viewing activation steering as a monolithic tool-flipping switch, \textsc{SAKIKO} treats it as one stage in an \textbf{adjudicated repair} process, progressing from directional error discovery and channel-keyed correction to destination verification and evidence-qualified licensing.

\section{The SAKIKO Protocol in Full}
\label{app:protocol}

The five stages named in Section~\ref{sec:framework} are the reader-facing
grouping; the protocol as executed runs eight, because \textit{Discovery} splits
into Discover and Support, \textit{Correction} into Estimate and Intervene, and
\textit{Verification} into Verify and Control. The six properties of the
evidence ladder (Appendix~\ref{app:licensing}) are what those stages establish,
not the stages themselves.

In order,
\[
\text{Discover} \rightarrow \text{Support} \rightarrow \text{Detect} \rightarrow
\text{Estimate} \rightarrow \text{Intervene} \rightarrow \text{Verify} \rightarrow
\text{Control} \rightarrow \text{Admit/Decline},
\]
of which the first seven are offline and only Detect and Intervene have an
inference-time counterpart (\S\ref{app:online}).

\subsection{Discover: channel construction}
\label{app:discover}

Baseline predictions are collected on the training split and grouped by the
ordered pair $(y^\star, \hat{y}_0)$, giving the full confusion topology rather than
a preselected set of transitions. Every off-diagonal cell is a candidate channel.
The number of channels is a property of the model and task and is not fixed: the
historical When2Call pipeline retains three channels on Phi-3.5 and Qwen2.5-7B,
whereas each sealed evaluation adjudicates a single preselected channel,
\texttt{cannot\_answer} $\rightarrow$ \texttt{tool\_call}, disclosed before sealing.

\subsection{Support: qualification before intervention}
\label{app:support}

A candidate channel advances only if it can be estimated on training data and
evaluated afterwards with enough rows to adjudicate. The frozen sealed criterion is
a minimum of $30$ channel errors in the evaluation population; the
development-stage gate applies a matching minimum of $30$ baseline-correct
reference rows per split, since a direction cannot be estimated without both
sides of the channel. Channels that fail
are set aside as \emph{unadjudicable}, not as uncorrectable: the stage reports a
property of the available sample, not of the model.

This gate is load-bearing rather than decorative. It stopped two settings in this
paper before any intervention was run. ACEBench produced a valid readout but no
directional transition met the preregistered per-split support minimum, and
Qwen3.5-9B failed that reference-side minimum with $29$ baseline-correct
rows against $30$, on a channel that had ample errors
(Appendix~\ref{app:declined}).

\subsection{Detect: observation site and Router}
\label{app:detect}

For a retained channel $c$, a per-channel Router is trained on the hidden state
$h_{\mathrm{obs}}$ read at a single observation layer $\ell_{\mathrm{obs}}$, taken at
the last prompt position of the residual stream. At inference it returns a channel,
a confidence, and a firing decision:
\begin{equation}
    r\!\left(h_{\mathrm{obs}}\right) \longrightarrow
    \bigl(\hat{c},\, p_{\hat{c}},\, \mathrm{fire}\bigr),
    \qquad
    \mathrm{fire} \iff p_{\hat{c}} \geq \tau_{\hat{c}} ,
    \label{eq:app-router}
\end{equation}
with the per-channel threshold $\tau_c$ selected on validation data and frozen with
the rest of the configuration. The Router is a firing rule, not a calibrated
probability: we do not recalibrate $p_{\hat{c}}$ and make no claim that it is
well calibrated in the sense of \citet{liu2024probecal}, whose recalibration of
internal tool-use probabilities addresses a different quantity. When the rule does not fire, the forward pass is left
untouched and $\hat{y}_1 = \hat{y}_0$.

\paragraph{Router: objective, populations and threshold.} The Router is a
per-channel linear discriminator fitted on training activations at
$\ell_{\mathrm{obs}}$: features are standardised to zero mean and unit variance,
and the classifier is $L_2$-regularised logistic regression at $C = 1.0$ with an
intercept, solved by \texttt{liblinear} to a tolerance of $10^{-4}$ under a cap of
$2{,}000$ iterations and a fixed seed, with non-convergence at that cap treated as
a hard failure rather than a fitted model. The two populations are asymmetric by
design. The positive class is the channel's own training errors, the rows whose
reference is $g$ and whose baseline prediction is $s$. The negative class is
\emph{every} baseline-correct training row, $\hat{y}_0 = y^\star$, and not only
those whose reference is $g$: the Router is asked to separate a channel-error
state from correct behaviour anywhere in the action space, which is the
discrimination its firing decision actually makes at inference. No class weighting
is applied, so the fit is unbalanced in the ratio the training split supplies.

The firing threshold is selected on a grid $\tau \in \{0.4, 0.5, 0.6, 0.7, 0.8\}$
evaluated on the development rows whose baseline prediction is the source action
--- the population the rule will face in operation, rather than the full
development split --- and $\tau_c$ is the smallest grid value at which precision
on that population reaches $0.50$. A channel is Router-eligible only if its
development ROC AUC against all-correct negatives is at least $0.75$ and the
selected threshold meets that precision floor; channels failing either are set
aside before any intervention. The fitted standardiser and coefficients are
written to the freeze with the selected $\tau$, so the operating point is fixed
before evaluation rather than tuned against it. The Router is not recalibrated as
a probability, and $p_{\hat{c}}$ is used only through this threshold.

\paragraph{What Router performance does and does not establish.} A Router that
separates channel-error states from a reference population establishes
\textsc{READABLE} and nothing further. In the layer analysis reported in the main
text, Router discrimination stays near-constant across a range of observation
layers while the correction direction estimated at those same layers degrades from
usable to noise. High Router AUC is therefore evidence that the state is decodable
at that site, not that intervening there will help.

\subsection{Estimate: correction directions}
\label{app:estimate}

Three estimators appear in this work. Only the first is used in the sealed
evaluations; the others are historical or serve as comparator arms.

\paragraph{Gradient / target-axis estimator ($d_{\mathrm{grad}}$).} The frozen
sealed estimator. In the Gemma configuration it is recorded as
\[
    d_{\mathrm{grad}} \;=\;
    \mathrm{unit}\!\left(\frac{1}{n}\sum_{i} \mathrm{unit}\!\left(
    G_{i,\text{gold}} - G_{i,\text{source}}\right)\right),
\]
a normalised mean of per-example normalised gradient differences between the gold
and source action scores, computed on training rows only
($n = 401$ training channel errors for Gemma). The direction is unit-norm and is
hashed into the freeze manifest before evaluation.

For one input and one candidate mode the scored quantity is the mean over that
candidate's token positions of the log-probability the model assigns to them. Its
gradient is taken with respect to the MLP forward output at $\ell_{\mathrm{inj}}$
--- the same tensor the intervention later modifies --- and reduced to a vector by
summing over sequence positions, which gives $G_{i,m}$. The inner normalisation
above makes the outer mean an average of directions rather than of magnitudes, so
no single large-gradient row dominates it. Nothing inside the estimator is
optimised, weighted, filtered or searched: there is no layer search, sign search,
sample selection or dose tuning. The difference-in-means comparator follows the
opposite subtraction by construction, mean correct-reference activation minus mean
channel-error activation, so that both directions point from the error state
toward the required action.

\paragraph{Difference-in-means (DiffMean).} The historical estimator, and in the
sealed protocol a comparator arm rather than the primary. It is the difference
between the mean activation of correct reference rows and the mean activation of
channel-error rows, computed on training data at a declared layer.

\paragraph{Principal component (PCA-1).} Used in historical configuration studies
where the difference-in-means direction was poorly aligned with the channel's
leading principal component. It is not part of the sealed protocol and is excluded
by construction from the formal arm inventory (Appendix~\ref{app:controls}).

\subsection{Intervene}
\label{app:intervene}

When the Router fires for channel $c$, the hidden state at the injection layer
$\ell_{\mathrm{inj}}$ is modified and the forward pass continues unchanged. The
site is the MLP forward output, after the MLP internals and before the decoder
residual addition, and the perturbation is applied at every sequence position,
matching the aggregation the direction was estimated under:
\begin{equation}
    h'_{\ell_{\mathrm{inj}}} \;=\; h_{\ell_{\mathrm{inj}}} \;+\; q_c\, s_c\, d_c ,
    \label{eq:app-intervention}
\end{equation}
where $d_c$ is the unit-norm correction direction, $s_c$ a scale statistic of the
activation norms at that site, and $q_c$ a relative dose. The product $q_c s_c$ is
the \emph{absolute perturbation budget}, and it is this quantity, not $q_c$, that is
held identical across every directional arm so that comparisons are budget-matched.
The $s_c$ above
is the median Euclidean norm of the MLP output at $\ell_{\mathrm{inj}}$, taken over
the training rows on which the committed Router fires. It is computed once from the
committed Routers and the baseline predictions, without a model load or any
development outcome, and enters the freeze as a number: the runner multiplies it by
the dose and verifies the achieved perturbation norm against the declared one
rather than recomputing the statistic at evaluation time.

For Qwen3-8B the budget is $q_c s_c = 41.5659848890$ at $q_c = 1.0$; for Gemma-2-9B
it is $0.15951178515982883$ at $q_c = 0.125$ with $s_c = 1.2760942812786307$. The
dose rule is \texttt{min(admissible)}: the smallest dose satisfying the declared
development criteria is taken, which makes the resulting placements conservative.

\paragraph{Observation and injection sites are distinct.} The layer at which the
state is read need not be the layer at which it is modified. In every evaluated
configuration the injection site \emph{precedes} the observation site
($\ell_{\mathrm{inj}} = 21$ against $\ell_{\mathrm{obs}} = 26$ for Qwen3-8B;
$24$ against $30$ for Gemma-2-9B), so a firing decision taken from
$h_{\mathrm{obs}}$ is realised by rerunning the input and intervening at
$\ell_{\mathrm{inj}}$ on a second forward pass. Model weights are frozen
throughout; nothing downstream of the injection site is modified.

\subsection{Offline construction versus online operation}
\label{app:online}

\begin{table}[t]
    \centering
    \small
    \setlength{\tabcolsep}{7pt}
    \renewcommand{\arraystretch}{1.18}
    \caption{Which stages require labelled data and which run at inference time.
    Only Detect and Intervene have an online counterpart; everything that
    consumes reference labels happens before deployment.}
    \label{tab:online}
    \begin{tabular}{@{}llll@{}}
        \toprule
        Stage & Phase & Requires labels & Data it may access \\
        \midrule
        Discover  & offline & yes & train split \\
        Support   & offline & yes & train, validation \\
        Detect    & offline (fit) / \textbf{online} (apply) & fit only & train, validation \\
        Estimate  & offline & yes & train only \\
        Intervene & offline (calibrate) / \textbf{online} (apply) & calibrate only & development \\
        Verify    & offline & yes & evaluation \\
        Control   & offline & yes & evaluation \\
        Admit/Decline & offline & yes & evaluation \\
        \bottomrule
    \end{tabular}
\end{table}

Table~\ref{tab:online} separates what is built offline from what runs at inference.
At inference the system holds a frozen backbone, one Router per retained channel, a
direction bank $\{d_c\}$ with doses $\{q_c\}$ and thresholds $\{\tau_c\}$, and a
hook at $\ell_{\mathrm{inj}}$. An input is scored, the Router either fires or does
not, and the answer is produced with or without one additive modification. No
weights are updated and no reference label is consulted.

\section{Licensing Criteria}
\label{app:licensing}

\subsection{The property hierarchy}

Six properties are adjudicated, and they are ordered only in the sense that the
later ones presuppose the earlier ones being posable:

\begin{description}\itemsep2pt
  \item[\textsc{ADJUDICABLE}] Reference labels, an error population, a reference
    population and a defined exposure set exist, and at least one action lies
    outside $\{g, s\}$ so that $O_c$ is not forced to zero by construction.
  \item[\textsc{READABLE}] Channel-error states are discriminable from a reference
    population at the declared observation site.
  \item[\textsc{STEERABLE}] The intervention has a direction-specific effect beyond
    zero, reversed, matched-random and wrong-site controls.
  \item[\textsc{CORRECTABLE}] Arrivals concentrate on $g$ rather than on other wrong
    actions. Undefined when $|\mathcal{Y}| = 2$.
  \item[\textsc{PRESERVING}] Collateral disruption stays within a prespecified
    bound. Two forms are distinguished throughout and only the first is
    adjudicated here. \emph{Population-level preservation} bounds $\mathrm{E1}$
    over all baseline-correct decisions, and is the property the frozen gate
    tests and the one a licence in this paper asserts. \emph{Exposure-conditional
    preservation} bounds $\mathrm{E2}$ over the decisions the Router actually
    touches; it is the stronger property, it is reported for every setting, and
    no setting in this paper attains it (Appendix~\ref{app:e2gate}).
  \item[\textsc{LICENSABLE}] The evidence meets a prespecified rule, at declared
    thresholds and confidence, for the claim being made.
\end{description}

A \emph{repair claim} is not a rung. It is the conjunction
\begin{equation}
    \textsc{REPAIR}(\pi) \;\equiv\;
    \textsc{CORRECTABLE}(\pi) \wedge \textsc{PRESERVING}_{\text{pop}}(\pi)
    \wedge \textsc{LICENSABLE}(\pi)
    \label{eq:app-repair}
\end{equation}
for a setting $\pi = (\text{model}, \text{dataset}, \text{channel},
\text{intervention}, \text{dose})$, and only a formal ADMIT licenses it.

\begin{table}[tb]
    \centering
    \small
    \caption{Correction across seven evaluated settings. \textit{Effect} is
    net gain on the locked test (historical) or the target gain on the evaluation
    population (sealed).
    \textit{Controls} summarises the comparison that decides
    direction specificity.
    Historical and sealed settings use different protocols and
    are not compared with each other. Historical settings carry no formal
    verdict: the frozen gate was written afterwards and was never run on them,
    and a \textsc{DECLINE} denotes insufficient evidentiary support under the
    protocol, not uncorrectability. Figure~\ref{fig:specificity},
    Appendix~\ref{app:historical} and Appendix~\ref{app:controls} give each
    setting's full control battery.
    $^\dagger$Secondary audit records report only
    the comparison with random directions.}
    \label{tab:correction_summary}
    \begin{tabular}{@{}llp{0.14\linewidth}p{0.30\linewidth}cl@{}}
        \toprule
        Model & Protocol & Effect & Controls & Specific & Verdict \\
        \midrule
        Phi-3.5-mini & historical & net +55 & random +14, reversed +14, wrong layer +3,
                       mismatched channel $-17$; five seeds +53 to +67 & yes & not sealed \\
        Qwen2.5-7B   & historical & net +79 & reversed +16; ten random directions,
                       mean +25.2, largest +50 & yes & not sealed \\
        Llama-3.1-8B & historical & ---$^\dagger$ & as many as 5 of 20 random
                       directions reach the real effect & \textbf{no} & not sealed \\
        Mistral-7B   & historical & net +12 & random mean +29, 17 of 20 reach the
                       real effect; reversed +19 & \textbf{no} & not sealed \\
        \midrule
        Qwen3-8B     & sealed & 0.276 & 0 of 59 random directions reach it
                       (largest 0.023) & yes & \textsc{ADMIT} \\
        Qwen3-4B     & sealed & 0.081 & 0 of 59 (largest 0.000) & yes & \textsc{DECLINE} \\
        Gemma-2-9B   & sealed & 0.073 & 0 of 59 (largest 0.000) & yes & \textsc{DECLINE} \\
        \bottomrule
    \end{tabular}
    \vspace{-1em}
\end{table}

\begin{table}[t]
    \centering
    \small
    \caption{The prospectively frozen SAKIKO evidence hierarchy. A formal repair claim is not a standalone property, but the strict conjunction $\textsc{REPAIR}(\pi) \equiv \textsc{CORRECTABLE}(\pi) \wedge \textsc{PRESERVING}_{\text{pop}}(\pi) \wedge \textsc{LICENSABLE}(\pi)$, whose preservation term is the population-level form.}
    \label{tab:ladder}
    \begin{tabular}{@{}lp{0.72\linewidth}@{}}
        \toprule
        Property Level & Operational Criterion \\
        \midrule
        \multicolumn{2}{@{}l}{\textit{Study Property}}\\
        \textsc{ADJUDICABLE} & Action space $|\mathcal{Y}| \ge 3$, with defined gold labels, baseline errors, and an exposed reference partition. \\
        \addlinespace[3pt]
        \multicolumn{2}{@{}l}{\textit{Representation Properties}}\\
        \textsc{READABLE}    & Error channel $c$ is linearly separable from reference states at observation layer $\ell_{\mathrm{obs}}$. \\
        \textsc{STEERABLE}   & Directional intervention outperforms zero-magnitude, matched-random, reversed
and layer-transposed controls, with cross-channel substitution added where the protocol records it. \\
        \addlinespace[3pt]
        \multicolumn{2}{@{}l}{\textit{Intervention Properties (Necessary for Repair)}}\\
        \textsc{CORRECTABLE} & Target-hit concentrates on the gold action rather than on other wrong actions. The frozen gate tests $\mathrm{TH}_c > 0.50$, a prespecified majority-of-exits criterion; it lies above the $1/(|\mathcal{Y}| - 1)$ uniform-alternative reference, which assumes an exit chooses evenly among the non-source actions and is not any model's measured behaviour. \\
        \textsc{PRESERVING}$_{\text{pop}}$ & Collateral disruption remains bounded. The frozen
  gate adjudicates this on the population denominator, $\mathrm{E1} \leq
  \epsilon_{\mathrm{tol}}$. The stronger exposure-conditional form,
  $\mathrm{E2} \leq \epsilon_{\mathrm{tol}}$, is reported throughout and is
  attained by no setting in this paper. \\
        \addlinespace[3pt]
        \multicolumn{2}{@{}l}{\textit{Evidential Property}}\\
        \textsc{LICENSABLE}  & Confidence intervals for $\mathrm{TH}_c$, $\mathrm{TG}_c$ and $\mathrm{E1}$ satisfy prospective thresholds under sample support. \\
        \bottomrule
    \end{tabular}
\end{table}

\subsection{The frozen ten-condition gate}

Each sealed setting is adjudicated by the ten conditions of
Table~\ref{tab:gatefull}, written in the Qwen3-8B preregistration before the first
sealed evaluation and applied unchanged to all three settings. The gate has a deliberate symmetry: every scientific quantity it
uses is tested twice, once at its point estimate and once at the edge of its
$95\%$ interval. Destination therefore contributes four conditions and preservation
two; the remaining four are integrity and specificity checks that are not about
effect size at all. Two of the four destination conditions are not independent at
the point estimate: since $\mathrm{TG}_c = (X_c / N_c)(2\,\mathrm{TH}_c - 1)$, when
$X_c > 0$ the conditions $\mathrm{TG}_c > 0$ and $\mathrm{TH}_c > 0.50$ are
algebraically equivalent. They are retained as separate numbered conditions because
the freeze numbered them so, and because the two quantities carry different
information --- $\mathrm{TG}_c$ retains coverage over $N_c$, $\mathrm{TH}_c$ the
destination composition over $X_c$ --- but they are not two independent pieces of
positive evidence, and their interval versions, Conditions 3 and 6, are not
equivalent.

\begin{table}[t]
    \centering
    \small
    \setlength{\tabcolsep}{5pt}
    \renewcommand{\arraystretch}{1.2}
    \caption{The canonical definition of a SAKIKO licence. Conditions are numbered
    as in the frozen artifacts and grouped here by what they check. An ADMIT
    requires all ten; the rule recorded in the freeze is \texttt{10/10; 9/10 =
    DECLINE}. Destination intervals are bootstrap intervals over channel errors
    with $10{,}000$ draws; the collateral interval is the one-sided upper
    bound from the same draws. Apart from the zero boundary in Conditions 2
    and 3, $0.50$ is the only destination threshold the gate uses.}
    \label{tab:gatefull}
    \begin{tabular}{@{}clp{0.17\linewidth}lll@{}}
        \toprule
        \# & Condition & Quantity & Population & Threshold & Layer \\
        \midrule
        \addlinespace[1pt]
        1  & channel support            & $N_c$                  & evaluation   & $\geq 30$    & integrity \\
        9  & zero arm exactness         & all endpoints          & routed       & exact match  & integrity \\
        10 & no structural failure      & run integrity          & ---          & none         & integrity \\
        \addlinespace[3pt]
        4  & specificity                & add-one $p$ over $K=59$ & $N_c$       & $\leq 0.05$  & specificity \\
        \addlinespace[3pt]
        2  & target gain, point         & $\mathrm{TG}_c$        & $N_c$        & $> 0$        & destination, point \\
        5  & target-hit, point          & $\mathrm{TH}_c$        & $X_c$        & $> 0.50$     & destination, point \\
        3  & target gain, interval      & $\mathrm{TG}_c$ CI low & $N_c$        & $> 0$        & destination, interval \\
        6  & target-hit, interval       & $\mathrm{TH}_c$ CI low & $X_c$        & $> 0.50$     & destination, interval \\
        \addlinespace[3pt]
        7  & collateral, point          & $\mathrm{E1}$          & $|C|$        & $\leq 0.05$  & preservation, point \\
        8  & collateral, interval       & $\mathrm{E1}$ CI upper & $|C|$        & $\leq 0.05$  & preservation, interval \\
        \bottomrule
    \end{tabular}
\end{table}

\subsection{What the licence adds over simpler rules}

\begin{figure}[t]
    \centering
    \includegraphics[width=\linewidth]{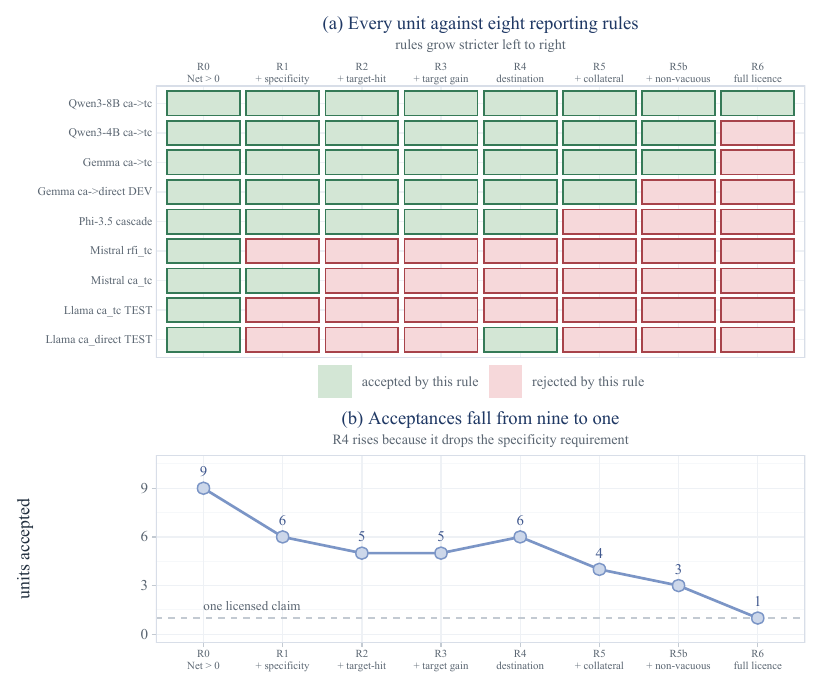}
    \caption{What the licence adds over simpler reporting rules. \textbf{(a)} Every
    evaluated unit against eight alternative reporting rules, applied
    retrospectively; the sequence is not nested, since R4 drops specificity rather
    than adding a requirement, and passing a rule here is a diagnostic reading
    rather than a formal verdict. Units span both
    protocol generations and include one development-stage setting
    (\textsc{Gemma ca$\rightarrow$direct DEV}), which is retained here because it is
    the clearest case of a vacuous preservation bound: zero exposed correct
    decisions. \textbf{(b)} How many units each rule accepts. A rule based on
    aggregate gain alone accepts all nine; the full frozen conjunction accepts one.
    R4 rises because it drops the specificity requirement rather than because the
    evidence improves.}
    \label{fig:licence-vs-rules}
\end{figure}

A natural objection is that the ten conditions are more apparatus than the evidence
requires: perhaps a simpler reporting rule would reach the same conclusions.
Figure~\ref{fig:licence-vs-rules} answers this directly by replaying every evaluated
unit --- sealed, historical and development alike --- against eight alternative
reporting rules, from ``Net gain above zero'' to the full frozen conjunction. The
sequence is not a nested tightening: R4 drops the specificity requirement rather
than adding to it. The replay is a retrospective diagnostic over units gathered
under different protocols, not nine independent confirmatory experiments, and a
unit passing a rule here is not a formal verdict for that unit.

A rule based on aggregate gain alone accepts all nine units. Adding specificity
removes three. Adding destination correctness on point estimates removes one more.
Requiring collateral to be bounded removes another, and requiring the collateral
bound to be non-vacuous removes a further one. The full conjunction accepts one.

Two features of the sequence are worth stating plainly. It is not monotone: R4 rises
to six because it tests destination \emph{without} requiring specificity, so it
readmits a setting whose movement is not established as direction-specific. And the
gap between R5b and R6 is the entire contribution of the interval layer --- three
units clear every point estimate and every non-vacuous bound, and two of them are
still declined.

\subsection{What a preservation-certifying gate would require}
\label{app:e2gate}

The gate this paper ran adjudicates preservation on $\mathrm{E1}$, and the reason
is historical rather than principled: the Qwen3-8B preregistration fixed that
denominator and the exposure-conditional field entered the record only with the
later Qwen3-4B run. We did not retrofit $\mathrm{E2}$ into an already-frozen
conjunction, so the licence reported here is population-level. We state here what
a gate that certified the property as defined would require, so that the gap is
specified rather than merely disclosed.

Two conditions would replace Conditions 7 and 8: a one-sided $95\%$ upper bound on
$\mathrm{E2}$ below $\epsilon_{\mathrm{tol}} = 0.05$ --- one-sided rather than
bootstrap, because the bootstrap bound is degenerate whenever no break is observed
(Appendix~\ref{app:stats}) --- and a minimum exposure $|C_{\mathrm{exp}}|$ large
enough for that bound to be attainable. The second is not a free parameter:
Appendix~\ref{app:stats} fixes it at $59$ exposed decisions with zero observed
breaks, $93$ with one, $124$ with two and $153$ with three. Against that
requirement Qwen3-8B exposes $6$, Gemma-2-9B $11$ and Qwen3-4B $50$, so none of
the three would clear the exposure minimum and the ADMIT would become a fourth
decline. No re-analysis of the existing records closes a gap of that size; it
requires a Router that exposes more decisions, or an evaluation population large
enough to supply them.

Two further conditions are absent from the frozen conjunction and belong in any
successor to it. The first is coverage: $\mathrm{TH}_c$ is computed over source
exits, so it says nothing about how much of the channel was reached, and a gate
written on it alone rewards a Router that fires rarely on easy rows. A successor
should condition on Router recall over $N_c$ and on gold arrivals over $N_c$
rather than over $X_c$; on the three sealed settings those arrival shares are
$0.437$, $0.298$ and $0.177$. The second is effect size: Condition 2 asks only
$\mathrm{TG}_c > 0$, which licenses an arbitrarily small positive gain provided
its interval clears zero, where a threshold tied to what a deployment would find
worth the collateral risk would be a stronger rule than a sign test. We add
neither here: both would have to be fixed before evaluation to mean anything, and
the evaluation is already spent.

\subsection{What each verdict asserts}

An ADMIT licenses the repair claim for that setting, under that protocol and dose,
and nothing more general. It is not a deployment-safety claim: the preservation
conditions the gate uses are computed on E1, the population denominator, and a
setting may pass them while leaving the exposure-conditional rate unbounded
(Appendix~\ref{app:q8b}).

A DECLINE states that the evidence is insufficient for the stronger claim under the
evaluated protocol and the available sample support. It does not state that the
intervention cannot work, that the setting cannot be corrected, or that one model
ranks below another. Because every property is conditional on the model, dataset,
channel, protocol and dose, a licence attaches to evidence gathered under a
declared protocol rather than to a model.

\section{Datasets and Action Spaces}
\label{app:data}

Three benchmarks appear in this work, in three different roles
(Table~\ref{tab:datasets}). Only one carries formal licence outcomes, and we keep the roles separate rather than pooling them
into a single breadth claim.

\subsection{When2Call}

The primary benchmark. Each input offers four candidate responses, one per
pre-execution action mode. Three of the four occur as reference labels; answering
directly appears only as a distractor. Labels are synthetically generated and
automatically assigned by the benchmark's authors, and we did not re-annotate them
(see Appendix~\ref{app:declined} for what this limits). A decision is read out by
teacher-forced scoring of the four complete candidates, and the predicted action is
the highest-scoring one; the readout is frozen and deterministic, so a baseline run
and an intervened run differ only through the intervention. The $3{,}652$ items are
split $2{,}556$ / $548$ / $548$ into training, validation and locked test.

In the released artifacts the four modes appear as \texttt{tool\_call},
\texttt{request\_for\_info}, \texttt{cannot\_answer} and \texttt{direct}; the paper
writes the last as \texttt{direct\_answer} for readability.

\subsection{MetaTool}

Used in its binary form for one model, where the only choice is whether to call a
tool. Because leaving the source mode is arriving at the target by construction,
$O_c = 0$ and target-hit is identically one: the destination question of
Appendix~\ref{app:notation} cannot be posed. We use MetaTool only for the argument
that the correction effect is not a single global shift in tool-call propensity,
since two opposite channels on disjoint populations both improve.

\subsection{ACEBench}

Instantiated as a second action ontology (\texttt{tool\_call}, \texttt{ask\_user},
\texttt{flag\_param\_error}, \texttt{cannot\_comply}), read out by free generation
and a rule parser rather than by teacher-forced scoring, and evaluated bilingually.
A criterion set locked before any result was seen first certified the readout, and
it passed: over $800$ items, accuracy $0.756$, macro-F1 $0.687$, no predicted class
above a $0.70$ share, $7.25\%$ unparseable, $100\%$ agreement on a replay check and
$0.98\%$ parser mislabelling on an audited subset of $102$ rows.

What failed was the channel structure, not the instrument. No directional transition
met the preregistered per-split support minimum, and a paraphrase-agreement check
required before channel discovery returned $0.56$ against a threshold of $0.80$.
Because $136$ of the $137$ native-label errors trace to the official parse rules,
the failure localises to the benchmark's size and wording sensitivity. Under the
locked decision rule this retires ACEBench from the intervention line: we report no
correction, no destination and no licence outcome for it. Two qualifications belong
with that verdict --- the parser figure is a point estimate on $102$ rows whose
stratification is undocumented, and the paraphrase reformulation is recorded in our
own audit as carrying a weaker format directive than the original, so the $0.56$ is
not clean evidence of wording fragility.

\begin{table}[t]
    \centering
    \small
    \setlength{\tabcolsep}{5pt}
    \renewcommand{\arraystretch}{1.2}
    \caption{Benchmark roles. Only When2Call carries formal licence outcomes.
    \emph{Gold} is whether gold labels are recoverable, \emph{Dest.} whether the
    destination question is posable, \emph{Interv.} whether an intervention was
    run. A dataset entering the table is not thereby evidence of breadth; the
    \emph{status} column records what each one actually supports.}
    \label{tab:datasets}
    \footnotesize
    \setlength{\tabcolsep}{4pt}
    \begin{tabular}{@{}lccccp{0.32\linewidth}@{}}
        \toprule
        Benchmark & Actions & Gold & Dest. & Interv. & Status \\
        \midrule
        \addlinespace[1pt]
        When2Call & 4 (3 gold) & yes & yes & yes & primary; all formal outcomes \\
        MetaTool  & 2          & yes & \textbf{no} & yes & historical; breadth of effect \\
        ACEBench  & 4          & yes & yes & \textbf{no} & framework transfer; support NO-GO \\
        \bottomrule
    \end{tabular}
\end{table}

\section{Models and Implementation}
\label{app:models}

\begin{table}[t]
    \centering
    \small
    \setlength{\tabcolsep}{4pt}
    \renewcommand{\arraystretch}{1.2}
    \caption{Evaluated settings and their frozen configuration. The horizontal rule
    separates the two protocol generations, which are never pooled. Entries marked
    \textsc{N/V} have no retained artifact to verify them against: the historical
    settings predate the configuration-freeze policy that governs the sealed rows,
    and we leave them unstated rather than quote an unverified value.}
    \label{tab:models}
    \begin{tabular}{@{}llccccc@{}}
        \toprule
        Model & Protocol & $\ell_{\mathrm{obs}}$ & $\ell_{\mathrm{inj}}$ & Estimator & Dose $q$ & Channels \\
        \midrule
        \addlinespace[1pt]
        Phi-3.5-mini & historical & 18 & 14 / 16$^{\ddagger}$ & DiffMean / PCA-1 & $\alpha = 10.0$ & 3 \\
        Qwen2.5-7B   & historical & 20 & 16 & DiffMean / PCA-1 & \textsc{N/V} & 3 \\
        Llama-3.1-8B & historical & \textsc{N/V} & \textsc{N/V} & \textsc{N/V} & \textsc{N/V} & \textsc{N/V} \\
        Mistral-7B-v0.3 & historical & \textsc{N/V} & \textsc{N/V} & \textsc{N/V} & \textsc{N/V} & \textsc{N/V} \\
        \midrule
        Qwen3-8B     & sealed & 26 & 21 & $d_{\mathrm{grad}}$ & 1.0   & 1 \\
        Qwen3-4B     & sealed & 26 & 21 & $d_{\mathrm{grad}}$ & 0.25  & 1 \\
        Gemma-2-9B   & sealed & 30 & 24 & $d_{\mathrm{grad}}$ & 0.125 & 1 \\
        \bottomrule
    \end{tabular}
\end{table}

\paragraph{Layer mapping.} Sealed observation and injection sites are derived from a
single normalised-depth rule anchored on the Qwen2.5-7B configuration,
$(\ell_{\mathrm{obs}}, \ell_{\mathrm{inj}}) = (20, 16)$ on $28$ layers:
$\ell^{\mathrm{new}} = \mathrm{round}\!\left(\ell^{\mathrm{old}} /
(L^{\mathrm{old}} - 1) \times (L^{\mathrm{new}} - 1)\right)$. This reproduces the
committed sites for Qwen3-8B ($26 / 21$) and yields $30 / 24$ for Gemma-2-9B. The
rule was fixed before the sealed settings were configured, so the sites are
mechanically derived rather than searched per model. Qwen3-4B shares Qwen3-8B's
layer count and therefore its sites, $26 / 21$, confirmed in its formal run log.

\paragraph{Dose grid.} Relative doses are drawn from a frozen six-point grid,
$q \in \{0.0,\, 0.125,\, 0.25,\, 0.5,\, 1.0,\, 2.0\}$, under the rule
\texttt{min(admissible)}. The three sealed settings take $1.0$, $0.25$ and $0.125$
from this grid; the absolute budgets are $q_c s_c = 41.5659848890$, $6.1748905405$
and $0.15951178515982883$ respectively. The historical Phi-3.5 configuration uses a
different parameterisation, a Router threshold $\tau = 0.4$ with $\alpha = 10.0$ at
seed $42$; two of its five seeds share that configuration and three re-select both
values on validation data.

$^{\ddagger}$Phi-3.5 records two injection variants, \texttt{mlp\_all} at $L14$ and
\texttt{mlp\_prompt} at $L16$. Its direction-estimation layer is recorded
inconsistently across two internal documents and is marked for verification below.

\paragraph{Execution environment.} The formal Qwen3-8B run is frozen under protocol
version \texttt{QWEN3\_STAGE2\_FORMAL\_V1} with \texttt{bfloat16} weights, eager
attention, batch size $1$, a single process and a single model load, fixed arm
order and fixed sample order, no checkpointing between scientific arms and no
resume between arms. A zero-magnitude gate runs before any endpoint is computed.
The recorded software stack for the ACEBench readout, which shares the environment,
is PyTorch $2.1.2{+}$cu121, Transformers $4.49.0$, CUDA $12.1$ on an NVIDIA
RTX~4090~D. Qwen3-8B is pinned in the preregistration to \texttt{Qwen/Qwen3-8B} at revision
\texttt{b968826d9c\allowbreak 46dd6066d1\allowbreak 09eabc6255\allowbreak 188de91218}, Qwen3-4B to \texttt{Qwen/Qwen3-4B} at revision
\texttt{1cfa9a7208\allowbreak 9121264592\allowbreak 14e8b04321\allowbreak 603b3df60c}, and Llama-3.1-8B to
\texttt{meta-llama/Llama-3.1-8B-Instruct} at revision
\texttt{0e9e39f249\allowbreak a16976918f\allowbreak 6564b8830b\allowbreak c894c89659}, \texttt{bfloat16}, seed $42$. Table~\ref{tab:models} carries no revision column, and the remaining model
revisions are not listed in this paper; they must be read from the repository
manifests.

\section{Formal Evaluation: Qwen3-8B}
\label{app:q8b}

\subsection{Procedure in temporal order}

Baseline predictions were collected and the confusion topology enumerated; the
channel \texttt{cannot\_answer} $\rightarrow$ \texttt{tool\_call} was selected and
the selection disclosed; the direction was estimated on training rows only and
hashed; the Router and threshold were fitted on training and validation; the dose
was calibrated on development data under the \texttt{min(admissible)} rule; the
runner, configuration, direction hashes and random seeds were frozen and a
gate-only pass verified them without opening the evaluation payload; an access
marker was written; the formal run executed once; endpoints were computed after the
zero-arm exactness gate passed.

\subsection{Destination-resolved outcome}

Table~\ref{tab:q8b-flow} resolves every decision in the two adjudicated cohorts, the $87$ channel errors and the $211$ baseline-correct rows. Of the
$87$ channel errors, $72$ were routed and $15$ were returned unchanged; of the
$72$, $20$ retained the source, $38$ reached the required action and $14$ landed
on a third. The Router exposed $6$ of the $211$ baseline-correct decisions and
broke none.

\begin{table}[t]
    \centering
    \small
    \setlength{\tabcolsep}{7pt}
    \renewcommand{\arraystretch}{1.2}
    \caption{Qwen3-8B formal run, every decision in the two adjudicated cohorts resolved; the remaining evaluation rows are neither channel errors nor baseline-correct. Of $87$ channel
    errors, $72$ were routed; the $15$ unrouted errors are returned unchanged by
    construction. Note the two Net conventions: \emph{whole} counts every fixed
    decision in the evaluation population, \emph{channel} counts arrivals in the
    adjudicated channel net of breaks, which is why the score-space comparator's
    $37$ arrivals and $2$ breaks appear as $+35$ in
    Table~\ref{tab:destination-compare}. The paper quotes both and never substitutes one for
    the other.}
    \label{tab:q8b-flow}
    \begin{tabular}{@{}llrl@{}}
        \toprule
        Baseline state & Final destination & Count & Population \\
        \midrule
        \addlinespace[1pt]
        channel error & source retained (routed)   & 20  & routed, 72 \\
        channel error & \textsc{GOLD ARRIVAL}      & \textbf{38}  & routed, 72 \\
        channel error & \textsc{OTHER WRONG}       & \textbf{14}  & routed, 72 \\
        channel error & unrouted, returned unchanged & 15 & $N_c = 87$ \\
        \addlinespace[4pt]
        baseline correct & retained correct & 6 exposed, 0 broken & $C_{\mathrm{exp}} = 6$ \\
        baseline correct & not exposed      & 205 & $|C| = 211$ \\
        \midrule
        \addlinespace[2pt]
        \multicolumn{2}{@{}l}{source exits $X_c$} & 52 & \\
        \multicolumn{2}{@{}l}{target-hit $\mathrm{TH}_c = 38/52$} & 0.7308 & CI $[0.60416, 0.84615]$ \\
        \multicolumn{2}{@{}l}{target gain $\mathrm{TG}_c = 24/87$} & 0.2759 & CI $[0.115, 0.425]$ \\
        \multicolumn{2}{@{}l}{Fixed / Broke / Net (whole)} & 40 / 0 / $+40$ & \\
        \multicolumn{2}{@{}l}{Net (channel-level)} & $+38$ & \\
        \multicolumn{2}{@{}l}{collateral E1} & 0 / 211 & upper bound $0.0141$ \\
        \multicolumn{2}{@{}l}{collateral E2} & 0 / 6 & upper bound $0.3930$ \\
        \bottomrule
    \end{tabular}
\end{table}

\subsection{Adjudication}

All ten conditions passed and the frozen verdict is
\texttt{FORMAL\_CONFIRMATORY\_SUCCESS}. None of the $59$ budget-matched random
directions reached the real direction's target gain; the largest random target gain
was $0.0230$ against the real $0.2759$, giving an add-one $p$ of $0.0167$. The zero
arm reproduced the baseline exactly, the reversed direction moved at most one
channel error off its source, and the same direction injected at $\ell = 26$ reached
a target gain of $0.0575$.

\paragraph{The licence carries a qualifier.} Preservation passes on E1 --- zero
breaks among all $211$ baseline-correct decisions, one-sided upper bound $0.0141$
--- but the Router exposed only six correct decisions, so the exposure-conditional
bound is $0.3930$ and E2 is \emph{vacuous} as evidence of preservation. The
statement this setting supports is that no collateral damage was observed in the
population, not that the intervention is safe on the decisions it touches. This
qualifier belongs to the claim and is not separable from it.

\section{Formal Evaluation: Qwen3-4B}
\label{app:q4b}

This setting is a formal adjudication, not a failed experiment. The intervention
produced directional movement that survived every specificity control, and every
point estimate cleared its threshold. The gate declined it on the precision of the
destination evidence alone.

\subsection{Outcome}

Of $124$ channel errors, $118$ were routed. The intervention produced $64$ source
exits, of which $37$ reached the required action and $27$ landed on a third wrong
action, giving $\mathrm{TH}_c = 0.5781$ over exits and $\mathrm{TG}_c = 10/124 =
0.081$ over all channel errors. Fixed and Broke were $40$ and $1$, a Net of $+39$ on
the whole population and $+36$ at channel level. The Router exposed $50$ correct
decisions and broke one of them, so E2 is $1/50$ with a one-sided upper bound of
$0.0914$; the population rate E1 passed both its point and interval conditions. The frozen
collateral audit quantifies why that pass is weak evidence: $164$ of the $214$
rows in the frozen denominator, $76.6\%$, were never perturbed at all, so the
gate's own metric is diluted by the rows the Router declined. The audit records
the exposure-conditional rate beside it as a required diagnostic disclosure
rather than as the frozen metric, and carries a claim lock stating that passing
the formal collateral endpoint must not be read as demonstrated deployment
safety. That lock is the reason this paper reports both denominators.
None of the $59$ budget-matched random directions reached the real target gain ---
the largest was $0.0000$ --- for an add-one $p$ of $0.0167$.

The named control arms separate cleanly here, and we report them because the same
battery on Qwen3-8B is quoted only in prose. Against the real arm's
$\mathrm{TG}_c = 0.081$ over $64$ exits, the zero arm reproduced the baseline exactly
and the reversed direction moved \emph{no} decision off its source at all ($0$
exits). Both misallocations are negative rather than merely weaker: the frozen
DiffMean estimator reaches $\mathrm{TG}_c = -0.016$ on $4$ exits, and the wrong-layer
arm $-0.024$ on $15$. Removing the Router leaves arrivals unchanged at $37$ but
raises breaks from $1$ to $6$, so gating buys a six-fold reduction in collateral at
no cost in arrivals. The score-space comparator is the one arm not dominated, and it is the
clearest illustration in the paper of why the two axes must be read together: it
reaches a higher target gain ($0.105$, $44$ arrivals) and breaks $6$ exposed
decisions against the activation arm's $1$. Neither arm is better on both axes at
once, which is precisely the situation an aggregate would collapse and a
destination-resolved account keeps visible. We report the contrast and claim no
ordering between them.

\subsection{Why the protocol declined}

\begin{table}[t]
    \centering
    \small
    \setlength{\tabcolsep}{5pt}
    \renewcommand{\arraystretch}{1.2}
    \caption{The two sealed Qwen3 settings against the frozen conjunction. Both
    clear every point estimate and both satisfy the frozen preservation conditions,
    which are written on $\mathrm{E1}$; on $\mathrm{E2}$ the point estimates are below
    the bound but the interval evidence is not ($0.3930$ and $0.0914$).
    The verdicts diverge on the two destination-interval conditions and on nothing
    else. Aggregate gain does not separate them: $+40$ against $+39$.}
    \label{tab:sealed-compare}
    \begin{tabular}{@{}clllcc@{}}
        \toprule
        \# & Condition & Layer & Threshold & Qwen3-8B & Qwen3-4B \\
        \midrule
        \addlinespace[1pt]
        1  & channel support        & integrity   & $\geq 30$   & pass & pass \\
        9  & zero arm exactness     & integrity   & exact       & pass & pass \\
        10 & no structural failure  & integrity   & ---         & pass & pass \\
        4  & specificity, $K=59$    & specificity & $p \leq 0.05$ & pass & pass \\
        2  & target gain, point     & destination & $> 0$       & pass & pass \\
        5  & target-hit, point      & destination & $> 0.50$    & pass & pass \\
        3  & target gain, interval  & destination & CI low $> 0$      & pass & \textbf{fail} \\
        6  & target-hit, interval   & destination & CI low $> 0.50$   & pass & \textbf{fail} \\
        7  & collateral, point      & preservation & $\leq 0.05$ & pass & pass \\
        8  & collateral, interval   & preservation & $\leq 0.05$ & pass & pass \\
        \midrule
        \addlinespace[2pt]
        \multicolumn{4}{@{}l}{\textit{target-hit point estimate}} & 0.7308 & 0.5781 \\
        \multicolumn{4}{@{}l}{\textit{target-hit 95\% interval}}  & $[0.604, 0.846]$ & $[0.456, 0.697]$ \\
        \multicolumn{4}{@{}l}{\textit{source exits}}              & 52 & 64 \\
        \multicolumn{4}{@{}l}{\textit{Net (whole)}}               & $+40$ & $+39$ \\
        \multicolumn{4}{@{}l}{\textbf{frozen verdict}}            & \textbf{ADMIT} & \textbf{DECLINE} \\
        \bottomrule
    \end{tabular}
\end{table}

Table~\ref{tab:sealed-compare} sets the two sealed Qwen3 settings against the
conjunction condition by condition. The target-hit interval $[0.4559, 0.6970]$ reaches below one half and the
target-gain interval includes zero. Under the frozen rule --- ten of ten admits,
nine of ten declines --- the protocol returned
\texttt{QWEN3\_4B\_FORMAL\_DECLINE}.

This outcome is the clearest demonstration in the paper that the licensing rule
changes the scientific conclusion. A reporting standard based on aggregate gain
would not distinguish these two settings at all: $+40$ against $+39$, a difference
of one decision. A standard that added specificity would still admit both, since
both clear the $K=59$ null. What separates them is the width of the destination
evidence at the sample sizes available, and Appendix~\ref{app:stats} shows that the
declines are what those sample sizes predict.

\section{Historical Experiments}
\label{app:historical}

\begin{figure}[t]
    \centering
    \includegraphics[width=\linewidth]{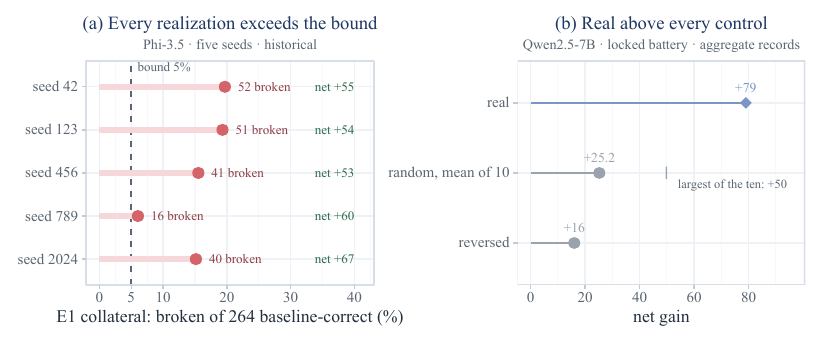}
    \caption{The historical record behind Sections~\ref{sec:results}
    and~\ref{sec:correction_to_repair}. \textbf{(a)} E1 collateral for each of the
    five artifact-backed Phi-3.5 seeds against the $5\%$ bound, with each run's Net
    gain: breakage ranges from $16$ to $52$ of $264$ baseline-correct decisions while
    the Net gain stays between $+53$ and $+67$, and every realisation exceeds the
    bound. \textbf{(b)} The Qwen2.5-7B locked battery: the real direction ($+79$)
    against its reversal ($+16$) and the mean of ten budget-matched random directions
    ($+25.2$, the largest reaching $+50$). Both panels are historical protocol and
    both survive as aggregate counts except where noted; they carry no formal
    verdict.}
    \label{fig:historical}
\end{figure}

The four historical settings developed the framework and are reported under the
protocol generation that produced them; Figure~\ref{fig:historical} shows the two
whose control batteries survive in full. They receive no formal verdict: the sealed
gate was written afterwards and was never run on them. Applying its criteria to
their recorded counts is a retrospective reading, and we label it as such wherever
it appears.

\subsection{Phi-3.5-mini}
\label{app:phi}

The only historical setting with surviving row-level records, and therefore the only
one whose destination composition can be resolved. Under the locked configuration at
seed $42$ the Router fired on $293$ decisions, comprising $200$ channel errors and
$93$ baseline-correct decisions. Of the $200$ errors, $47$ retained their source and
$153$ left it; of those exits, $107$ reached the required action and $46$ landed on a
third wrong action, for $\mathrm{TH} = 107/153 = 0.6993$. Of the $93$ exposed correct
decisions, $41$ survived and $52$ were broken.

The aggregate for the same run is Fixed $107$, Broke $52$, Net $+55$. Collateral is
$52/93 = 55.9\%$ on the exposure-conditional denominator and $52/264 = 19.7\%$ on the
population denominator; the one-sided upper bound on the former is $0.6468$. Both
exceed the $5\%$ bound the sealed protocol later prespecified.

\paragraph{Seed variation.} Repeating the full pipeline under five seeds gives Net
gains of $55$, $54$, $53$, $60$ and $67$, positive in all five, with Broke ranging
from $16$ to $52$. This is stability across the evaluated seeds rather than a general
property: two of the five share the frozen configuration and three re-select their
threshold and dose on validation data, so the runs are configuration variants rather
than repeated draws from one configuration. An independent execution of the same
code, seed and configuration fired on $243$ decisions rather than $293$ and broke
$44$, for reasons the surviving records cannot establish. The magnitude of the damage
depends on the execution; its presence does not.

\paragraph{Controls.} Against the real configuration's $+55$: a random direction
gives $+14$, the reversed direction $+14$, injection at a different layer $+3$, and
one channel's direction applied to another channel $-17$, which is worse than
leaving the model alone. Four of these five arms survive only as aggregate counts,
so the battery establishes an ordering rather than a decomposition of the effect.

\subsection{Qwen2.5-7B}
\label{app:qwen25}

Applying the frozen channel configurations to the locked test set, $92$ baseline
errors become correct while $13$ already-correct decisions become wrong, a Net of
$+79$. Three channels are retained --- a follow-up question answered instead with a
tool call, a refusal answered instead with a tool call, and a refusal answered
instead with a direct answer --- and together they carry about $78\%$ of all baseline
errors.

Its destination composition \emph{cannot} be adjudicated: the When2Call records for
this setting survive only as aggregate counts, so $A_c$ and $O_c$ cannot be
separated. This is why Qwen2.5-7B appears on the hierarchy as passing specificity
and then stopping, rather than as a destination result.

Against its own control battery the real direction gives $+79$, the reversed
direction $+16$, and ten budget-matched random directions a mean of $+25.2$ with a
largest value of $+50$. Across five seed-level runs the real direction again exceeds
every random draw, although three of those runs drew a single random direction rather
than ten --- a heterogeneity in null strength that we record rather than average over.

\subsection{MetaTool}
\label{app:metatool}

The same machinery was applied to two opposite channels, unnecessary tool calls and
missed ones, on disjoint sample populations. Both improved, with mean Net gains of
$8.6$ and $13.0$ across five runs and $21.6$ when applied together, against a
baseline accuracy of $0.773$.

The argument this supports is narrow and specific: because the two channels move
tool-calling behaviour in \emph{opposite} directions on non-overlapping inputs, a
single global shift in tool-call propensity cannot account for the result. It
supports nothing about destination, because in a binary action space $O_c = 0$ by
construction.

\subsection{Mistral-7B and Llama-3.1-8B}
\label{app:negatives}

Both settings fail the specificity comparison, and the framework stops there. On
Llama-3.1-8B as many as $5$ of $20$ random directions reach the real effect. On
Mistral-7B the real direction's Net gain of $+12$ falls \emph{below} the random mean
of $+29$, $17$ of $20$ random directions reach it, and the reversed direction gives
$+19$.

Two qualifications are required. Both results rest on secondary audit records rather
than on primary row-level artifacts, and both are flagged in those records as
estimator-suspect and compared against $20$ random directions rather than the $59$ of
the sealed protocol. The conclusion these settings support is therefore that, under
the evaluated configuration, estimator and site, there is insufficient evidence that
the correction is caused by the recovered direction. They do not establish that no
correctable direction exists in these models.

Llama-3.1-8B is, descriptively, the clearest case of movement without arrival in our
records, with $19$ exits reaching the required action against $44$ landing elsewhere.
Because the setting fails specificity first, it cannot serve as evidence that a
\emph{specific} correction fails to arrive, and we do not use it as such.

\section{Control Experiments}
\label{app:controls}

Every directional arm in a sealed evaluation receives the identical absolute
perturbation budget $q_c s_c$ and passes through the identical Router gate as the
real arm; Table~\ref{tab:controls} states what each arm exists to rule out. Matching the budget rather than the relative dose is what makes the
comparison meaningful across layers and models.

\begin{table}[t]
    \centering
    \small
    \setlength{\tabcolsep}{5pt}
    \renewcommand{\arraystretch}{1.2}
    \caption{Control arms in the sealed protocol, transcribed from the frozen
    control specification. The \emph{alternative removed} column states what each
    arm exists to rule out. Budget and gate are identical to the real arm unless
    noted. Qwen3-8B values are quoted; Gemma-2-9B uses the same inventory at
    $\ell_{\mathrm{inj}} = 24$ with wrong-layer at $\ell = 30$.}
    \label{tab:controls}
    \footnotesize
    \setlength{\tabcolsep}{4pt}
    \begin{tabular}{@{}lp{0.19\linewidth}p{0.17\linewidth}p{0.34\linewidth}@{}}
        \toprule
        Arm & Budget / site & Population & Alternative removed \\
        \midrule
        \addlinespace[1pt]
        baseline & none & all 548 rows & --- defines both populations \\
        zero & none & routed & that endpoints are computed on a mis-specified pipeline; must reproduce the baseline exactly before any endpoint is read \\
        \addlinespace[2pt]
        real $d_{\mathrm{grad}}$ & $q_c s_c$ at $\ell_{\mathrm{inj}}$ & routed & --- primary arm \\
        \addlinespace[2pt]
        $K = 59$ fresh randoms & $q_c s_c$ at $\ell_{\mathrm{inj}}$ & routed & that any perturbation of this magnitude would move the decision; this is the primary null \\
        reverse $-d_{\mathrm{grad}}$ & $q_c s_c$ at $\ell_{\mathrm{inj}}$ & routed & that the axis matters but its sign does not \\
        wrong layer & \emph{same} $q_c s_c$ at $\ell_{\mathrm{obs}}$ & routed & that the direction works anywhere in the stack; the budget is \emph{not} renormalised to the other layer's activation norm, which would change site and budget together \\
        frozen DiffMean & $q_c s_c$ at $\ell_{\mathrm{inj}}$ & routed & that the specific estimator is irrelevant \\
        ungated & $q_c s_c$ at $\ell_{\mathrm{inj}}$ & all rows with pred $=$ source & that the Router contributes nothing \\
        score-space & frozen $b_c$, no model arm & routed & that the effect requires an internal intervention at all; a reducibility test rather than a specificity control \\
        \bottomrule
    \end{tabular}
\end{table}

\paragraph{The random null is near-orthogonal to what it tests.} The $59$
directions are verified unit-norm to $2.7 \times 10^{-9}$, mutually
near-orthogonal (median $|\cos| = 0.011$, maximum $0.061$) and near-orthogonal to
the calibrated direction (maximum $|\cos| = 0.050$), so the null is a
budget-matched comparison against directions that do not overlap the one under
test. Seeds are disjoint from development and the direction matrix is hash-pinned
in the freeze manifest before the evaluation split is opened.

The construction is recorded in full and is reproducible from the manifest alone:
for each $i < K$ the seed is the first eight bytes of $\mathrm{SHA256}(\text{salt}
\,\|\, i)$, a \texttt{PCG64DXSM} generator draws $z \sim \mathcal{N}(0, I)$ in
float64, the vector is normalised in float64 and cast once to float32. The null is
therefore isotropic by construction rather than by inspection. The manifest also
records, per vector, the seed input, the seed digest and the vector digest, so any
reader can regenerate the set and check it against the frozen hashes. Four
exclusions are asserted in the same record and are what make the null a null: no
rejection sampling, no orthogonalisation, no cosine filtering and no outcome
matching, with no regeneration and no post-hoc selection. The cosines quoted above
were computed \emph{after} the complete set was frozen, and no vector was removed
on their basis.

\paragraph{Arms excluded by construction.} The frozen inventory explicitly excludes
a same-layer DiffMean at the injection site, another channel's direction, an
orthogonalised mismatch, any pooled or shared direction, PCA or rank-2 directions,
any new estimator, and any additional dose. Excluding them before sealing is what
prevents the control battery from becoming a search.

\paragraph{Gating.} Removing the gate changes the outcome, but not uniformly, and we
describe it only by its measured effects. On Phi-3.5 an ungated arm reverses the sign
of the effect within an independent execution of the locked configuration, from
$+62$ to $-32$; on both MetaTool channels the ungated arms fall to $-39$ and $-34$;
on Qwen3-8B the ungated Net drops from $+38$ to $+13$ while its collateral rises
from $0/6$ to $29/176 = 16.5\%$. On Gemma-2-9B the ungated arm is a genuine
counterexample: its Net is not lower but marginally higher ($+16$ gated against
$+17$ ungated), and its collateral \emph{rate} is lower rather than higher
($4/188 = 2.1\%$ ungated against $1/11 = 9.1\%$ gated), though a single break on
eleven exposed rows makes that rate comparison unstable, which is why we state the
counterexample on Net where it is unambiguous. The five settings with both a gated
and an ungated arm are Phi-3.5, the two MetaTool channels taken together,
Qwen2.5-7B, Qwen3-8B and Gemma-2-9B, and Net is the only endpoint recorded for
every one of them. Read on Net, gating raises the effect in three of the five ---
Phi-3.5, MetaTool and Qwen3-8B --- and lowers it in two, on Qwen2.5-7B by $+79$
against $+94$ and on Gemma-2-9B by $+16$ against $+17$. We therefore make no
necessity claim on Net. Arrivals per broken decision and collateral rate
order the arms differently in places, and we do not substitute one for another. On Qwen2.5-7B the ungated
arm reaches a \emph{larger} Net ($+94$ against $+79$) by acting on $293$ decisions
instead of $202$ and absorbing more damage. What gating raises consistently, in the
three settings where both arms are directly comparable, is the number of arrivals per
broken decision: from $5.27$ to $7.08$ on Qwen2.5-7B, from $5.25$ to $17.00$ on
Gemma-2-9B, and from $1.45$ on Qwen3-8B to a gated arm that breaks nothing.

\paragraph{The named arms on Gemma-2-9B.} The third sealed setting separates in
the same direction as Qwen3-4B and we give it for completeness. Against the real
arm's $\mathrm{TG}_c = 0.073$ over $27$ exits ($17$ arrivals, $10$ elsewhere, $1$
break of $11$ exposed), the zero arm reproduced the baseline exactly and the
reversed direction again moved \emph{no} decision off its source. Both
misallocations are negative: the frozen DiffMean estimator produced a single exit
and no arrival at all ($\mathrm{TH}_c = 0$, $\mathrm{TG}_c = -0.010$), and the
wrong-layer arm $-0.010$ on $3$ exits. The ungated arm reaches $21$ arrivals
against the gated $17$ but breaks $4$ exposed decisions against $1$, which is the
$5.25$ to $17.00$ change in arrivals per break quoted above.

\paragraph{The comparator does not order consistently.} Across the three sealed
settings the activation arm and the frozen score-space comparator admit no stable
ordering. On Qwen3-8B the activation arm leads on target gain ($0.276$ against
$0.184$) and on Gemma-2-9B it leads again ($0.073$ against $0.052$), while on
Qwen3-4B the comparator leads ($0.105$ against $0.081$). The comparator's $b_c$ is
calibrated on development data at the selected dose and never tuned on the sealed
split. That the sign of the difference changes with the setting, while no paired
test survives multiplicity correction, is why the frozen interpretation rule for
this contrast is honoured rather than set aside: we report it and claim no
ordering.

\section{Destination-Resolved Evaluation}
\label{app:destination}

Net is not a wrong statistic. It is the right statistic for the question ``did
accuracy improve'', and on that question it is exact. The claim of this paper is
narrower: Net is a many-to-one compression of the five-class flow in
\eqref{eq:app-partition}, and the classes it discards are the ones a repair
claim depends on.

\subsection{Two interventions, equal aggregate, different destinations}

On the sealed Qwen3-8B population the preregistered activation intervention and the
score-space comparator produce nearly identical aggregates and materially different
destination compositions.

\begin{table}[t]
    \centering
    \small
    \setlength{\tabcolsep}{7pt}
    \renewcommand{\arraystretch}{1.2}
    \caption{Activation intervention against the score-space comparator on the same
    channel and population. The aggregates are close; the destinations are not. The
    paired tests are not significant after multiplicity correction and the comparison
    was not prespecified, so this contrast is \emph{descriptive}: it establishes
    neither that activation intervention is better nor that the two differ in
    general.}
    \label{tab:destination-compare}
    \begin{tabular}{@{}lcc@{}}
        \toprule
        Quantity & Activation ($d_{\mathrm{grad}}$) & Score-space \\
        \midrule
        \addlinespace[1pt]
        gold arrivals $A_c$        & 38 & 37 \\
        other wrong $O_c$          & 14 & 21 \\
        source exits $X_c$         & 52 & 58 \\
        target-hit $\mathrm{TH}_c$ & 0.7308 & 0.6379 \\
        target gain $\mathrm{TG}_c$ & 0.2759 & 0.1839 \\
        \addlinespace[3pt]
        Fixed / Broke (whole)      & 40 / 0 & 41 / 2 \\
        Net (whole)                & $+40$ & $+39$ \\
        Net (channel-level)        & $+38$ & $+35$ \\
        \bottomrule
    \end{tabular}
\end{table}

A study reporting only the aggregate would have treated these two interventions as
interchangeable: $+40$ against $+39$. Resolved by destination they differ by seven
decisions redistributed to a third wrong action. The paired comparisons are exact
McNemar $p = 1.0$ for arrivals and $p = 0.189$ for redistributions, and the archived
artifact records that all-548 paired correctness was not prespecified as a paired
comparison. We therefore report the contrast and draw no superiority claim in either
direction.

\section{Mechanistic Insights and Stage Decoupling}
\label{sec:mechanism}

\textbf{Decoupled Dynamics Across Probing, Geometry, Gating, and Dosage.}
The analyses in this subsection are a preregistered follow-up and are exploratory: they are not part of the frozen conjunction, and they locate no component. Mechanistic analyses on Qwen2.5-7B and prospective cohorts demonstrate sharp dissociations between representation decoding and intervention efficacy. Across layers 16--22 of Qwen2.5-7B, linear probe decodability plateaus at $\text{AUROC} \approx 0.96$, yet steering vector norms scale non-linearly ($2.12 \rightarrow 14.26$) and validation net gain jumps from $+13$ at layer 16 to $+46$ at layer 20, where layer 16 steering fails due to near-zero PC1 alignment. Across geometric formulations, PCA extraction outperforms difference-in-means ($+50$ vs.\ $+36$ locked net gain), while cross-channel injections produce severe negative performance ($-17$ net gain on Phi-3.5-mini); the channels are related but not interchangeable, with pairwise cosines of $0.488$, $0.569$ and $0.706$ between Qwen2.5-7B's three When2Call directions measured at a common layer. Ablating router gating triggers catastrophic degradation on unconstrained runs, causing net gains to flip negative on Phi-3.5-mini ($+62 \rightarrow -32$) and MetaTool ($-39$ and $-34$); conversely, gating sharpens precision, reducing breaks from 6 to 1 on Qwen3-4B while holding arrivals constant at 37. Read on Net, gating is not uniformly necessary: it raises Net on Phi-3.5, MetaTool and Qwen3-8B, and lowers it on Qwen2.5-7B ($+79$ gated against $+94$ ungated) and marginally on Gemma-2-9B ($+16$ against $+17$), so the effect on Net is setting-dependent. What gating changes consistently is exposure and the collateral that follows from it. Finally, dose-response curves on Qwen3-4B reveal non-monotonic scaling, where target-hit rates span $0.525$ to $0.911$ across dosage regimes, peaking at intermediate injection magnitudes before regressing.

\textbf{The Multi-Dimensional Architecture of Latent Steering.}
These mechanistic investigations reveal that representation decodability does not entail causal steerability: the internal layers that maximally linearly separate error states are not necessarily the computational sites where interventions effectively shift downstream decisions. Furthermore, the sensitivity of outcomes to extraction geometry and injection dosage demonstrates that causal steering acts within specific, multi-dimensional subspaces rather than along arbitrary, monolithic rank-one directions. Finally, gating limits exposure but does not by itself establish preservation among exposed decisions: it shields the decisions the Router declines, and the evidence here says nothing stronger about the ones it touches. Consequently, mechanistic repair cannot be treated as a static model property, but must be conceptualized as a delicate balance between extraction geometry, injection dosage, and exposure control.

\section{Synthesis: The Spectrum from Movement to Repair}
\label{subsec:synthesis}

\textbf{Systematic Stage-Wise Attrition Across Model Architectures.}
Mapping the complete empirical evaluation across \textsc{SAKIKO}'s staged hierarchy demonstrates progressive attrition, wherein individual model configurations fail at distinct, predictable criteria (Table~\ref{tab:correction_summary}). In the historical cohort, Llama-3.1-8B and Mistral-7B fail the foundational specificity controls, failing to differentiate from budget-matched random vectors. Moving higher in the hierarchy, Phi-3.5-mini clears destination correctness ($\mathrm{TH}_c = 0.699$) but fails preservation by corrupting 52 baseline-correct decisions, while Qwen2.5-7B exhibits high net gain ($+79$) but lacks row-level destination logging. In the sealed prospective cohort, Qwen3-4B and Gemma-2-9B satisfy point thresholds for correctability and preservation, but fail the licensing stage due to inconclusive uncertainty intervals crossing null boundaries. Ultimately, only Qwen3-8B successfully navigates the entire five-stage pipeline, clearing all structural, empirical, and statistical thresholds to earn an \textsc{ADMIT} verdict. The historical stops carry no formal verdict: the gate was written afterwards and was never run on them.

\textbf{Enforcing the Boundary Between Behavioral Steering and Verified Repair.}
This progressive attrition shows that behavioral movement, destination-resolved correction, clean-state preservation, and statistical certitude do not inherently coincide in the settings we evaluate. Standard steering literature routinely conflates these levels, claiming repair whenever an internal intervention yields positive aggregate deltas or forces an agent out of an error state. By establishing an evidence pipeline where each verification stage addresses independent failure modes, \textsc{SAKIKO} exposes how aggregate gains can conceal lateral error redistribution, collateral damage, and finite-sample uncertainty. Formalizing this staged adjudication transforms internal representation interventions from ad-hoc activation steering into a rigorous, verifiable methodology for agentic repair.
\section{Statistical Procedures}
\label{app:stats}

\paragraph{Bootstrap intervals.} Destination intervals are percentile bootstrap
intervals resampling \emph{channel errors} --- not exits, and not evaluation rows ---
with $10{,}000$ draws. Resampling the error population rather than the exits keeps
the denominator of $\mathrm{TG}_c$ fixed and propagates the uncertainty in how many
decisions leave their source at all. Collateral is bounded two ways and we keep
them distinct. Gate Condition 8 uses the bootstrap $95\%$ upper bound on E1, as
written in the preregistration. The bounds quoted in the tables are one-sided
Clopper--Pearson upper bounds at $95\%$, which at these counts are the more
conservative of the two --- $0/211$ gives $0.0000$ under the bootstrap and
$0.0141$ under Clopper--Pearson --- so every collateral bound the paper prints is
at least as strict as the one the gate applied.

\paragraph{With no observed break the bootstrap bound is degenerate.} This is not a
cosmetic difference in the one case it decides. When $B = 0$ every nonparametric
resample of the correct population also has $B = 0$, so the bootstrap $95\%$ upper
bound on E1 is exactly zero and carries no finite-sample information: Condition 8
cannot fail on a setting with no observed break, however few decisions were exposed.
Qwen3-8B is that setting. We state the consequence rather than leave it to be
inferred --- its Condition 8 pass is uninformative as written, and the bound a reader
should use is the one-sided Clopper--Pearson value, $0.0141$ on $0/211$. Substituting
Clopper--Pearson into the gate does not change the verdict, since $0.0141 \leq 0.05$,
so no result here turns on the choice; but a gate written now should name the
rare-event-informative bound prospectively, and Appendix~\ref{app:e2gate} states what
else it would have to require.

\paragraph{Undefined target-hit under resampling.} A resample in which $X_c = 0$ leaves
$\mathrm{TH}_c$ undefined. At the observed exit shares the probability of drawing one
is below $10^{-13}$ in every sealed setting --- $1.7 \times 10^{-14}$ for Gemma-2-9B,
the least favourable of the three --- so the case does not arise in $10{,}000$ draws
and no discard rule was exercised.

\paragraph{Empirical random null.} Specificity is judged by an add-one Monte Carlo
$p$-value over $K$ budget-matched random directions passed through the same Router
gate. Here and below $K$ counts random directions; the cardinality of the action
space, written $K$-way in the title, is $|\mathcal{Y}|$ throughout the formal text.
The null is an empirical distribution over isotropic directions at a matched
budget, not a permutation of labels:
\begin{equation}
    p_{\text{add-one}}
    \;=\;
    \frac{1 + \#\{\, T_{\mathrm{rand}} \geq T_{\mathrm{real}} \,\}}{K + 1},
    \qquad T = \mathrm{TG}_c .
    \label{eq:app-addone}
\end{equation}
With $K = 59$ and no random direction reaching the real target gain, this attains its
floor of $1/60 = 0.0167$ in each sealed setting. The value $K = 59$ was chosen on
computational cost alone, and it bounds the resolution of the test: $0.0167$ is the
smallest $p$ this design can report, not a measure of how far the real direction
exceeds the null. The observed margins are reported separately --- the largest random
target gain was $0.0230$ on Qwen3-8B and $0.0000$ on both Qwen3-4B and Gemma-2-9B.

\paragraph{Design sensitivity.} The main text reports how many source exits a true
target-hit requires before a one-sided exact binomial test against $0.50$ rejects at
$\alpha = 0.05$ with $80\%$ power. Table~\ref{tab:power} gives the full curve.

\begin{table}[h]
    \centering
    \small
    \setlength{\tabcolsep}{8pt}
    \renewcommand{\arraystretch}{1.15}
    \caption{Source exits required to certify a true target-hit above $0.50$. This
    approximates the gate's bootstrap criterion rather than replacing it, and we
    apply it only to the settings in this paper.}
    \label{tab:power}
    \begin{tabular}{@{}lcccccc@{}}
        \toprule
        true target-hit & 0.73 & 0.68 & 0.63 & 0.60 & 0.58 & 0.55 \\
        \midrule
        source exits needed & 30 & 49 & 93 & 158 & 245 & 620 \\
        \bottomrule
    \end{tabular}
\end{table}

Read at the observed point estimates, Qwen3-8B's $52$ exits exceed its requirement of
about $30$, while Gemma-2-9B's $27$ and Qwen3-4B's $64$ fall several times short of
theirs. The declines are what the sample sizes predict.

\paragraph{Preservation power.} Certifying a $5\%$ collateral bound at $\alpha = 0.05$
with zero observed breaks requires at least $59$ exposed decisions: $0/58$ gives
$0.0503$ and only $0/59$ reaches $0.0495$. With one break it requires $n \geq 93$,
with two $n \geq 124$, with three $n \geq 153$. No setting in this paper attains the
required exposure, and no re-analysis, denominator substitution or pooling can close
a gap of that size on the existing records.

\paragraph{Boundary conventions.} Target-hit is undefined when $X_c = 0$ and is
reported as such rather than as zero or one. A collateral rate computed on zero
exposed rows is recorded as \emph{vacuous} and never as preservation. Multiplicity:
the paired activation-versus-score-space tests are corrected and reported as
non-significant; no other comparison in the sealed protocol is multiple.

\section{Sealing, Preregistration and Provenance}
\label{app:provenance}

\subsection{Separation of development and confirmation}

TRAIN and DEV data were used to select the channel, estimator, observation and
injection sites, dose and Router threshold. The evaluation population was sealed
throughout that process. Before any formal run, a gate-only pass verified --- without
opening the evaluation payload --- all artifact hashes, model and tokenizer identity,
software versions, the selected channel and configuration, direction hashes, formal
random count and hashes, seed disjointness, arm completeness, fixed arm order, batch
size, the single-load path, that the zero gate precedes endpoint computation, the raw
record schema, the primary/secondary hierarchy, the bootstrap specification, the
support-outcome branch, the VOID rules, the sealed-evaluation firewall, and an empty
formal output namespace.

\paragraph{Rows excluded for hardware, and where they came from.} Six of the
$3{,}652$ rows could not be run: the four-mode gradient stack in bfloat16 with
eager attention exceeds $24$\,GB at their sequence lengths. Three fall in train
and three in development; \emph{none} falls in the sealed evaluation split, and
the exclusion manifest lists their raw indices. The boundary is a natural gap in
the data rather than a chosen threshold --- the last feasible row has maximum
sequence length $3{,}169$ and the first infeasible one $4{,}572$ --- and it was
fixed before any endpoint was observed. A release scan over the committed record
files reports no credentials, no activation caches and no dataset payload, and
confirms the record schema: immutable sample identifiers, four-mode
log-probability scores, predictions, Router probabilities, arm identity, dose and
destination, with no prompt, candidate string or generated text retained.

An access marker was then written, and the formal run executed \emph{once}. For
Gemma-2-9B the one-shot ledger records \texttt{formal\_run\_count = 1},
\texttt{sealed\_rows\_read = 548} and $65$ executed arms ($6$ named forward arms plus
$59$ randoms), against authorised repository head
\texttt{23d6da2a64\allowbreak 639ca50d43\allowbreak d94de04a5b\allowbreak 6d6275d896}.

\paragraph{What the freeze fixed, and what it forbade.} The preregistration is not
only a list of thresholds. It fixes, before any evaluation access, the primary
statistic, the ten-condition conjunction, an \emph{ordered} secondary hierarchy that
is explicitly non-promotable, the fourteen-step execution order, the bootstrap
method and its derived seed, the support rule, thirteen named \textsc{VOID}
conditions, and a set of interpretation rules that map each possible outcome onto
what may be claimed from it. Three of those provisions do work that a
post-hoc reader cannot replicate. The conjunction may not be replaced with a
best-looking subset after execution. Secondary results may not rescue a failed
primary. And the interpretation rules bind in both directions: the frozen rule for
the case in which the score-space comparator matches or exceeds the activation arm
states that no claim may then be made that activation intervention provides
behaviour unavailable to a frozen score shift. $K = 59$ was fixed from a
compute-cost table over $K \in \{39, 59, 79\}$ and declared independent of
development rankings, effect sizes, anticipated $p$-value and random geometry. The
runner reads every frozen quantity from committed artifacts at run time rather than
restating it in code, and the raw record schema retains the \emph{achieved}
perturbation norm alongside the intended one, so a drift between specification and
execution would be visible in the records rather than silent.

\paragraph{What selection conditions on.} The channel, estimator and dose were chosen
before sealing, on development data, and every sealed result conditions on that
disclosed selection. This is legitimate and it is not blind; we state it rather than
present the sealed evaluation as if the configuration had been arbitrary.

The frozen disclosure names the criteria and the rejected alternatives. The
adjudicated channel was retained because it cleared the support, Router, geometric
and dose gates, and because its collateral bound is a real constraint rather than a
formality: $368$ development rows are both routed-eligible and baseline-correct. Two
other eligible channels were excluded before sealing and neither is reported as
confirmatory afterwards, which would be selection after the fact. One,
\texttt{cannot\_answer} $\rightarrow$ \texttt{direct}, was excluded because its
collateral rate is \emph{structurally} zero --- the pinned split contains no gold
\texttt{direct} rows, so no routed row can ever be baseline-correct and its safety
gate carries no information. The other,
\texttt{request\_for\_info} $\rightarrow$ \texttt{tool\_call}, ranked the two
interventions in the opposite order on development data and was set aside on that
basis. The single confirmatory channel is therefore chosen on development evidence
that includes how the two compare, so the sealed run establishes destination-correct
repair on the channel it adjudicates and is not a general ranking of activation
intervention against a frozen score shift. That is the claim the paper makes, and
the disclosure is what keeps it that size.

\subsection{Corrections applied after the fact}

Table~\ref{tab:provenance} lists every historical value a later audit corrected,
with the artifact that is authoritative in each case.

\begin{table}[t]
    \centering
    \small
    \setlength{\tabcolsep}{5pt}
    \renewcommand{\arraystretch}{1.2}
    \caption{Historical values that later audits corrected. Each correction makes a
    derived index agree with the frozen artifact it cites; no frozen artifact was
    modified and no verdict changed.}
    \label{tab:provenance}
    \footnotesize
    \begin{tabular}{@{}p{0.19\linewidth}p{0.17\linewidth}p{0.17\linewidth}p{0.34\linewidth}@{}}
        \toprule
        Quantity & Superseded & Authoritative & Reason \\
        \midrule
        \addlinespace[1pt]
        Qwen3-4B target-hit CI lower & 0.4561 & \textbf{0.4559} & index carried an independent recomputation; the frozen verdict artifact is authoritative \\
        Qwen3-4B target-hit CI upper & 0.6984 & \textbf{0.6970} & same \\
        Gemma target-hit CI lower & 0.4400 & \textbf{0.4444} & moved by $0.0044$, beyond the Monte-Carlo noise the captions had allowed \\
        \addlinespace[3pt]
        Phi collateral denominator & $52/264$ as \emph{the} rate & \textbf{$52/93$} as the estimand & $264$ arises from using the channel-assignment field as the firing decision; the firing rule is \texttt{route $\neq$ none}, giving $293 = 200 + 93$ \\
        \addlinespace[3pt]
        Qwen3-4B, Gemma rung label & ``admitted at Level-1'' & \textbf{FORMAL DECLINE} & falsified by the primary verdict artifacts \\
        \addlinespace[3pt]
        ``Level-1 / Level-2'' & used as licence tiers & post-hoc descriptive only & no prospectively frozen definition exists in any artifact \\
        \bottomrule
    \end{tabular}
\end{table}

\paragraph{A resolved attribution.} A prose rendering of one mechanism table places
the layer-sweep observation --- difference-in-means norms of $2.12$ and $14.26$ with
validation gains of $+13$ and $+46$ --- in a column headed \emph{Phi-3.5}. The
machine-readable indices place it on \emph{Qwen2.5-7B}'s refusal-to-direct-answer
channel, and name the source file accordingly; the recorded sweep is the four-point
series $2.12$, $5.88$, $10.41$, $14.26$ over layers $16$ to $22$, with
$\cos(\text{DiffMean}, \mathrm{PC}_1) = 0.0002$ at $L16$. The tabular rendering is a
formatting error and the indices are authoritative. The main text and this appendix
attribute the sweep to Qwen2.5-7B.

The one-shot ledgers record what a restart was permitted to be. On Qwen3-4B a
mechanical \textsc{VOID} for incomplete all-arm execution was followed by exactly
one authorised restart, and the ledger states its terms: every arm re-executed
from scratch, no resume from partial records, and the voided attempt retained by
its identifier. Its gate-only pass is recorded as $25/25$ checks against the
frozen specification before any sealed row was read. One mechanical incident is
recorded in the Gemma formal freeze and its pre-repair runner is retained
alongside the incident note. Three pre-access gate repairs and one
mechanical VOID occurred, all before any endpoint was observed, and all are disclosed
with before-and-after hashes.

\section{Declined and Failed Settings}
\label{app:declined}

A decline is an output of the protocol, not a failure of it. That framing does not
convert a failure into a success, and each entry below states plainly what did not
hold.

\paragraph{Insufficient support: Qwen3.5-9B.} An eighth model was evaluated and never
reached the intervention stage. On every naive indicator it was the most promising
setting in the panel --- the highest baseline accuracy ($0.4533$), the least collapse
onto a single predicted class, and errors lying closest to the decision boundary. Its
accuracy gain is unevenly distributed across gold classes: recall on a follow-up
question roughly doubles relative to the other models ($0.326$ against $0.170$ and
$0.132$) while recall on a refusal falls to the lowest in the panel ($0.087$). Because
reference and error rows within a gold class compete for a fixed number of items, that
redistribution starves the reference side of every refusal channel --- including the
exact channel carrying the sealed evaluations --- leaving $29$ reference rows against
a frozen minimum of $30$, and enriches the one channel that three models already agree
is not linearly readable at the observation site. The sealed population was never
opened. The stop is split-sensitive: under $30$ retrospective group-preserving
reallocations the refusal channels were eligible in $56.7\%$ of them, against a median
achievable support of $34$.

\paragraph{Dataset ineligibility: ACEBench.} Described in Appendix~\ref{app:data}.
The readout passed every validity criterion; the channel structure did not meet the
preregistered support minimum, and the paraphrase gate returned $0.56$ against $0.80$.
Retired from the intervention line by the locked decision rule.

\paragraph{Readable but not direction-specific: Llama-3.1-8B, Mistral-7B.} Described
in Appendix~\ref{app:negatives}. Both fail the specificity comparison, both rest on
secondary audit records, and both used $K = 20$ rather than $K = 59$.

\paragraph{Destination evidence insufficient: Qwen3-4B, Gemma-2-9B.} Described in
Appendix~\ref{app:q4b}. Both clear every point estimate and both satisfy preservation
on the frozen $\mathrm{E1}$ denominator; both fail the two destination-interval
conditions and nothing else.

\paragraph{Preservation not certified anywhere, including the ADMIT.} No setting in
this paper attains exposure-conditional preservation certification. Qwen3-8B's
$0/6$ gives an upper bound of $0.3930$; Gemma-2-9B's $1/11$ gives $0.3644$;
Qwen3-4B's $1/50$ gives $0.0914$; Phi-3.5's $52/93$ gives $0.6468$. The paper reports
that none attains it, which is a true statement fully supported by the data, and does
not claim preservation certification anywhere.

\section{Reproducibility}
\label{app:repro}

\begin{table}[t]
    \centering
    \small
    \setlength{\tabcolsep}{5pt}
    \renewcommand{\arraystretch}{1.2}
    \caption{Where each result in the paper is recorded. Paths are relative to the
    repository root. Row-level records exist only where indicated; the remaining
    historical settings survive as aggregate counts and cannot be re-resolved by
    destination.}
    \label{tab:repro}
    \begin{tabular}{@{}llc@{}}
        \toprule
        Result & Artifact & Row-level \\
        \midrule
        \addlinespace[1pt]
        Qwen3-8B formal & \texttt{final/results/qwen3\_stage2\_formal/} & yes \\
        Qwen3-4B formal & \texttt{final/results/qwen3\_4b\_w2c\_formal\_v1/} & yes \\
        Gemma-2-9B formal & \texttt{.../gemma/formal\_freeze/} & yes \\
        Phi-3.5 multi-seed & \texttt{final/results/clean/p1\_multiseed\_table.csv} & seed 42 only \\
        Mistral-7B & \texttt{final/results/mistral7b\_w2c\_sakiko\_ca/} & no \\
        Llama-3.1-8B & \texttt{final/results/phase8\_prospective\_llama/} & no \\
        ACEBench readout & \texttt{final/results/acebench\_generation\_readout/} & yes \\
        Qwen3.5-9B stop & \texttt{research\_exploration/qwen35\_sakiko/} & n/a \\
        \addlinespace[3pt]
        Integrity recheck & \texttt{final\_freeze/GLOBAL\_INTEGRITY\_RECHECK.md} & --- \\
        Index corrections & \texttt{final\_evidence/EVIDENCE\_CORRECTIONS.md} & --- \\
        Standing constraints & \texttt{constraints/STANDING\_CONSTRAINTS.md} & --- \\
        \bottomrule
    \end{tabular}
\end{table}

Table~\ref{tab:repro} gives the artifact location for every result in the paper.
The formal Qwen3-8B run is reproduced by
\texttt{python scripts/qwen3\_stage2\_formal.py --run-formal} under protocol version
\texttt{QWEN3\_STAGE2\_FORMAL\_V1}, guarded by an approval environment variable; the
configuration check is \texttt{--gate-only} and runs without opening the evaluation
payload. Determinism controls are greedy decoding, batch size $1$, fixed arm order,
fixed sample order, a single process and a single model load, with no checkpointing
or resume between scientific arms.

All \texttt{*.jsonl}, \texttt{*.npz} and \texttt{*.npy} artifacts are tracked with
Git LFS. A checkout without \texttt{git lfs pull} yields pointer files of
approximately $132$ bytes that read as empty rather than failing, so any reproduction
attempt should verify artifact sizes before interpreting an empty result.

\end{document}